\documentclass[10pt,journal,compsoc]{IEEEtran}

\usepackage{epsfig}
\usepackage{epstopdf}
\usepackage{graphicx}
\usepackage{amsmath}
\usepackage{amssymb}
\usepackage{algorithm}
\usepackage{algorithmic}
\usepackage{multirow}
\usepackage{color,soul}
\usepackage[utf8]{inputenc}
\usepackage{mathtools}
\usepackage{enumerate}
\usepackage{comment}

\usepackage{color, colortbl}
 
\usepackage[colorinlistoftodos, shadow]{todonotes/todonotes}

\ifCLASSOPTIONcompsoc
  \usepackage[nocompress]{cite}
\else
  \usepackage{cite}
\fi

\ifCLASSINFOpdf
\else
\fi

\definecolor{LightGray}{gray}{0.9}
\definecolor{DarkGray}{gray}{0.8}

\newcommand{\average}[1]{%
    {\em{#1}}%
}

\graphicspath{
	{./figures/}
}

\begin{document}

\title{Deformable Parts Correlation Filters for Robust Visual Tracking}

\author{Alan Lukežič,
        Luka Čehovin,~\IEEEmembership{Member,~IEEE,}
        and~Matej Kristan,~\IEEEmembership{Member,~IEEE}
\IEEEcompsocitemizethanks{\IEEEcompsocthanksitem A. Lukežič, L. Čehovin and M. Kristan are with the Faculty of Computer and Information Science, University of Ljubljana, Slovenia.\protect\\
E-mail: see http://www.vicos.si/}
}

\markboth{Paper under revision}%
{Lukežič \MakeLowercase{\textit{et al.}}: Deformable Parts Correlation Filters for Robust Visual Tracking}

\IEEEtitleabstractindextext{%
\begin{abstract}
Deformable parts models show a great potential in tracking by principally addressing non-rigid object deformations and self occlusions, but according to recent benchmarks, they often lag behind the holistic approaches. The reason is that potentially large number of degrees of freedom have to be estimated for object localization and simplifications of the constellation topology are often assumed to make the inference tractable. We present a new formulation of the constellation model with correlation filters that treats the geometric and visual constraints within a single convex cost function and derive a highly efficient optimization for MAP inference of a fully-connected constellation. We propose a tracker that models the object at two levels of detail. The coarse level corresponds a root correlation filter and a novel color model for approximate object localization, while the mid-level representation is composed of the new deformable constellation of correlation filters that refine the object location. The resulting tracker is rigorously analyzed on a highly challenging OTB, VOT2014 and VOT2015 benchmarks, exhibits a state-of-the-art performance and runs in real-time.
\end{abstract}

}

\maketitle

\IEEEdisplaynontitleabstractindextext

%
\IEEEpeerreviewmaketitle



\section{Introduction} \label{sec:introduction}

Short-term single-object visual tracking has received a significant attention of the computer vision community over the last decade with numerous conceptually diverse tracking algorithms being proposed every year. Recently several papers reporting experimental comparison of trackers on a common testing ground have been published~\cite{otb_cvpr2010,alov_pami2014,kristan_vot2013,kristan_vot2014}. Results show that tracking quality depends highly on the expressiveness of the feature space in the object appearance model and the inference algorithm that converts the features into a presence score in the observed parameter space. Most of the popular trackers apply holistic appearance models which capture the object appearance by a single patch. In combination with efficient machine-learning and signal processing techniques from online classification and regression, these trackers exhibited top performance across all benchmarks~\cite{hare_struck, babenko_mil, grabner_oab, bolme2010visual}. Most of these approaches apply sliding windows for object localization, and some extend the local search in the scale space~\cite{danelljan2014accurate,samf_eccv2014,henriques2015tracking,comanichu_kernel_pami2003} to address the scale changes as well.
\begin{figure}[h!] 
\begin{center}
\includegraphics[width=0.9\linewidth]{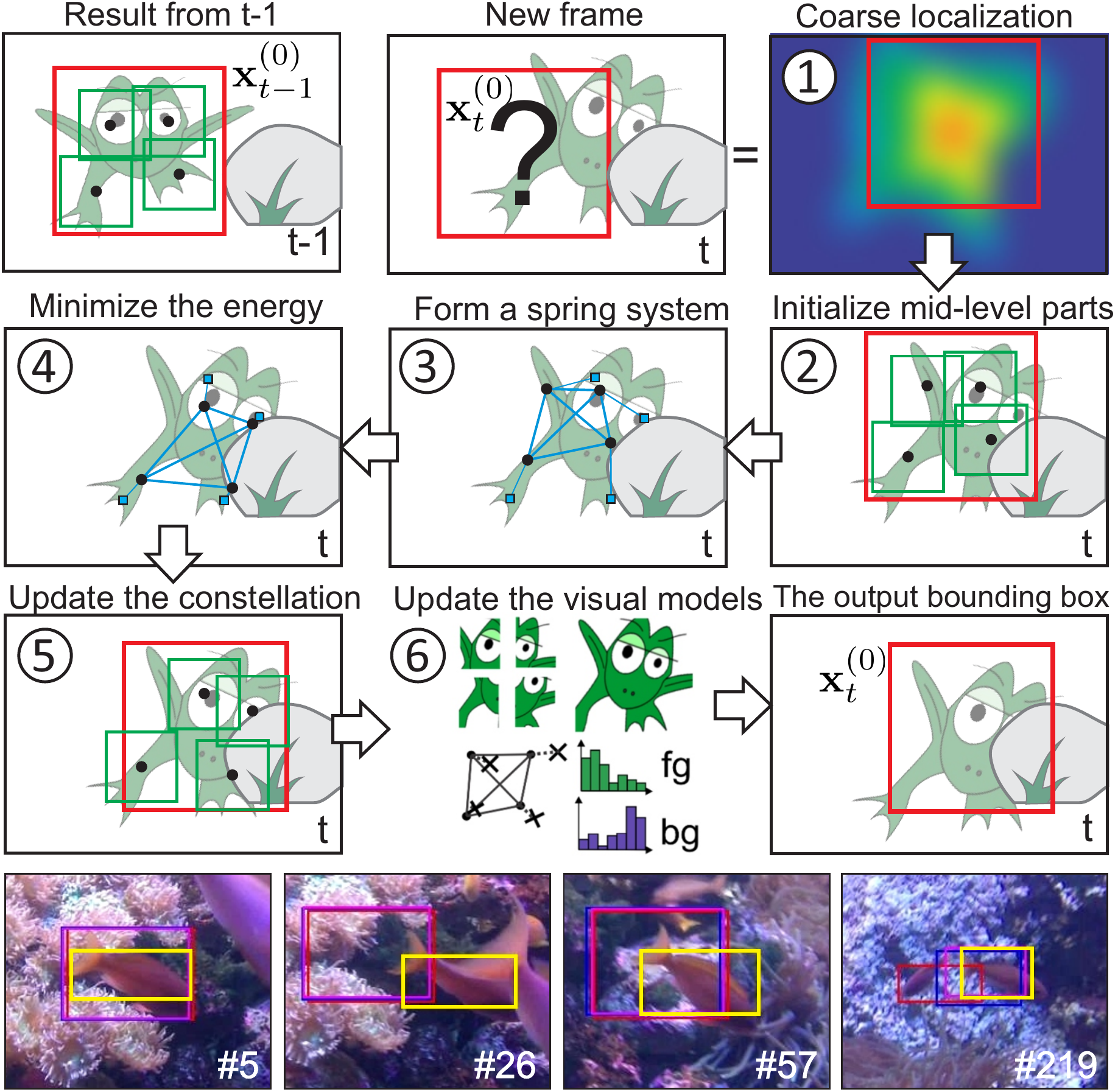}
\end{center}
   \caption{Illustration of coarse-to-fine tracking by spring system energy minimization in a deformable part model (top). Tracking examples with our tracker DPT (yellow), KCF (red), IVT (blue) and Struck (magenta) are shown in the bottom.}
\label{fig:overview}
\end{figure}

Nevertheless, a single patch often poorly approximates objects that undergo significant, potentially nonlinear, deformation, self occlusion and partial occlusions, leading to drift, model corruption and eventual failure. Such situations are conceptually better addressed by part-based models that decompose the object into a constellation of parts. This type of trackers shows a great potential in tracking non-rigid objects, but their performance often falls behind the holistic models~\cite{kristan_vot2014}, because of the large number of degrees of freedom that have to be estimated in the deformation model during tracking. Čehovin et al.~\cite{lgt_tpami2013} therefore propose that part-based models should be considered in a layered framework that decomposes the model into a global and local layer to increase the stability of deformation parameters estimation in presence of uncertain visual information. Most part-based trackers use very small parts, apply low-level features for the appearance models, e.g., histograms~\cite{lgt_tpami2013,kwon_tracking_sampling_tpami2014} or keypoints~\cite{Maresca2013,pr2011_artner} and increase their discrimination power by increasing the number of parts. Object is localized by optimizing a trade-off between the visual and geometric agreement. Most of the recent trackers use star-based topology, e.g.~\cite{kwon_tracking_sampling_tpami2014,dgt2014tip,ruiyao_partbased_cvpr2013,godec2013cviu,part_context_bmvc2014,Maresca2013}, or local connectivity, e.g.~\cite{lgt_tpami2013}, instead of a fully-connected constellation~\cite{pr2011_artner} to make the inference tractable, but at a cost of a reduced power of the geometric model.

In this paper we present a new class of layered part-based trackers that apply a geometrically constrained constellation of local correlation filters~\cite{bolme2010visual, henriques2015tracking} for object localization. We introduce a new formulation of the constellation model that allows efficient optimization of a fully-connected constellation and adds only a negligible overhead to the tracking speed. Our part-based correlation filter formulation is cast in a layered part-based tracking framework~\cite{lgt_tpami2013} that decomposes the target model into a coarse layer and a local layer. A novel segmentation-based coarse model is introduced as well. Our tracker explicitly addresses the nonrigid deformations and (self-)occlusions, resulting in increased robustness compared to the recently proposed holistic correlation filters~\cite{henriques2015tracking} as well as state-of-the-art part-based trackers.

\subsection{Related work} \label{sec:related_work}

Popular types of appearance models frequently used for tracking are generative holistic models like color histograms~\cite{Collins2005} and subspace-based~\cite{ross_ivt,zhang_ct} or sparse reconstruction templates~\cite{mei_tpami2011}. Several papers explored multiple generative model combinations~\cite{Collins2005, hong_multitask_view_iccv2013} and recently Gaussian process regressors were proposed for efficient updating of these models~\cite{gaoECCV2014}. The cost function in generative holistic models reflects the quality of global object reconstruction in the chosen feature space, making the trackers prone to drifting in presence of local or partial object appearance changes or whenever the object moves on a visually-similar background. This issue is better addressed by the discriminative trackers which train an online object/background classifier and apply it to object localization. Early work includes support vector machines (SVM)~\cite{avidan_svm_tracking}, online Adaboost~\cite{grabner_oab}, multiple-instance learning~\cite{babenko_mil} and recently excellent performance was demonstrated by structured SVMs~\cite{hare_struck}. A color-based discriminative model was recently presented in~\cite{posseger_color_cvpr15} that explicitly searches for potential visual distractors in the object vicinity and updates the model to increase the discriminative power. The recent revival of the matched filters~\cite{Naidu1974} in the context of visual tracking has shown that efficient discriminative trackers can be designed by online learning of a correlation filter that minimizes the signal-to-noise ratio cost function. These filters exhibit excellent performance at high speeds, since learning and matching is carried out by exploiting the efficiency of the fast Fourier transform. Bolme et al.~\cite{bolme2010visual} introduced the first successful online matched filter, now commonly known as a correlation filter tracker. Their tracker was based on grayscale templates, but recently the correlation filters have been extended to multidimensional features~\cite{henriques2015tracking,danelljan2014accurate,samf_eccv2014}, and Henriques et al.~\cite{henriques2015tracking} introduced kernelized versions. Scale adaptation of correlation filters was investigated by Danneljan et al.~\cite{danelljan2014accurate} and Zhang et al.~\cite{zhang_2015_robust} who applied correlation filters to the scale space and~\cite{li_2015_reliable} who combined votes of multiple automatically allocated filters. Zhang et al.~\cite{zhang_stc_eccv2014} have shown the connection to spatio-temporal context learning. Hong et al.~\cite{muster_cvpr15} have recently integrated correlation filters in a multi-store tracking framework and demonstrated excellent performance. In fact, the correlation filter-based trackers have demonstrated excellent performance across all the recent benchmarks. Still, these trackers suffer from the general drawbacks of holistic models is that they do not explicitly account for deformation, self occlusion and partial occlusions, leading to drift, model corruption and eventual failure. This issue is conceptually better addressed by models that decompose the object into parts.

The part-based trackers apply constellations of either generative or discriminative local models and vary significantly in the way they model the constellation geometry. Hoey~\cite{Hoey2006} used a flock-of-features tracking in which parts are independently tracked by optical flow. The flock is kept on object by identifying parts that deviate too far from the flock and replacing them with new ones. But because of weak geometric constraints, tracking is prone to drifting. Vojir et al.~\cite{vojir_fot_2014} addressed this issue by significantly constraining the extent of each part displacement and introduced tests of estimation quality. Tracking robustness is increased by only considering the part displacements deemed accurately estimated. Martinez et al.~\cite{Martinez2008} proposed connecting triplets of parts and tracked them by kernels while enforcing locally-affine deformations. The local connectivity resulted in inefficient optimization and parts required careful manual initialization. Artner et al.~\cite{pr2011_artner} proposed a key-point-based tracker with a fully-connected constellation. They use the geometric model that enforces preservation of inter-keypoint distance ratios. Because the ratios are not updated during tracking and due to the ad-hoc combination of geometric and appearance models, the resulting optimization is quite brittle, requiring manual initialization of parts and the resulting tracker handles only moderate locally-affine deformations. Pernici et al.~\cite{Pernici2013} address nonrigid deformations by oversampling key-points to construct multiple instance-models and use a similarity transform for matching. But, the tracker still fails at significant nonrigid deformations. Several works simplify a geometric model to a star-based topology in interest of simplified optimization. A number of these works apply part detectors and a generalized Hough transform for localization. Examples of part detectors are key-points~\cite{Maresca2013}, random forest classifiers~\cite{godec2013cviu}, ferns~\cite{yangreal} and pixels~\cite{Duffner2013}.
Cai et al.~\cite{dgt2014tip} apply superpixels as parts combined with segmentation for efficient tracking, but the high reliability on color results in significant failures during illumination changes. Kwon et al.~\cite{kwon_tracking_sampling_tpami2014} apply generative models in a star-based topology with adding and removing parts and {\v C}ehovin et al.~\cite{lgt_tpami2013} increase the power of the geometric model by local connectivity. Both approaches require efficient stochastic optimizers for inference. Yao et al.~\cite{ruiyao_partbased_cvpr2013} address the visual and geometric model within a single discriminative framework. They extend the structured SVM~\cite{hare_struck} to multiple part tracking, but cannot handle scale changes. This model was extended by Zhu et al.~\cite{part_context_bmvc2014} to account for context as well, but uses a star-based topology for making the inference tractable. Context was also used by Duan et al.~\cite{duan_2012_group} where tracking multiple objects or object parts was used to resolve ambiguities.
 
Part-based trackers often suffer from the potentially large number of parameters of the deformation model to be estimated from uncertain/noisy visual data. This is addressed by the layered paradigm of part-based trackers introduced by Čehovin et al.~\cite{lgt_tpami2013}. This paradigm decomposes the tracker architecture into a global coarse and a local appearance layer. The global layer contains coarse target representations such as holistic templates and global color histograms, while the local layer is the constellation of parts with simple local appearance description. The paradigm applies a top-down localization to gradually estimate the state parameters (i.e., target center and part locations) and bottom-up updates to update the appearance models. Čehovin et al.~\cite{lgt_tpami2013} analyzed various modalities used at the global layer (i.e., color, local motion and shape) and their influence on tracking. They have concluded that color plays the most important role at the scale of the entire object.

\subsection{Our approach and contributions} \label{sec:approach}
 
Our main contribution is a new class of fully-connected part-based correlation filter trackers. Most part-based trackers apply star-based topology to simplify the inference or combine geometrical and visual constraints in an ad-hoc fashion often leading to a nonconvex optimization problem. In contrast, our formulation treats the geometric and visual constraints within a single convex cost function. We show that this cost function has a dual formulation of a spring system and show that MAP inference of the constellation can be achieved by minimizing the energy of the dual spring system. We derive a highly efficient optimizer that in practice results in a very small computational overhead during tracking. 

The tracker is formulated within the theoretical framework of layered deformable parts~\cite{lgt_tpami2013} that decomposes the tracker into a coarse representation and a mid-level representation. The coarse representation is composed of a holistic correlation filter and a novel global color model. The mid-level representation is composed of local correlation filters fully-connected by the new constellation model. Tracking is performed by top-down localization and bottom-up updates (Figure~\ref{fig:overview}): The coarse model initializes the mid-level representation at approximate object location. An equivalent spring system is formed and optimized, yielding a MAP constellation estimate. The parts are updated and the estimated constellation is used to update the coarse model. In contrast to the standard holistic correlation filters, the proposed deformable parts tracker naturally addresses the object appearance changes resulting from scale change, nonrigid deformations and (self)occlusions increasing the tracking robustness.

Our tracker and the proposed constellation optimization are analyzed in depth. The tracker is rigorously compared against a set of state-of-the-art trackers on a highly challenging recent benchmarks OTB~\cite{otb_cvpr2010}, VOT2014~\cite{kristan_vot2014} and VOT2015~\cite{kristan_vot2015} and exhibits a state-of-the-art performance. Additional tests show that improvements come from the fully-connected constellation and the top-down/bottom-up combination of the coarse representation with the proposed deformable parts model.

\section{Deformable parts tracker}\label{sec:methods}

As it is a common practice in visual tracking, the tracker output at time-step $t$ is an axis-aligned bounding box. In our case this region is estimated by the deformable parts correlation filter as we describe in this Section. Our tracker is composed of a coarse representation described in Section~\ref{sec:coarseRepresent}, and of a deformable constellation of parts, a mid-level object representation described in Section~\ref{sec:midLevelRepresent}. In the following, we will denote the part positions by $(\cdot)^{(i)}$, where the index $i=0$ denotes the root part in the coarse layer and indexes $i>0$ denote parts in the constellation. Since both representations apply kernelized correlation filters (KCF)~\cite{henriques2015tracking} for part localization, we start by briefly describing the KCF in Section~\ref{sec:kcf}.
 
\subsection{Kernelized correlation filters}\label{sec:kcf}

\begin{figure}[h]
\begin{center}
\includegraphics[width=0.8\linewidth]{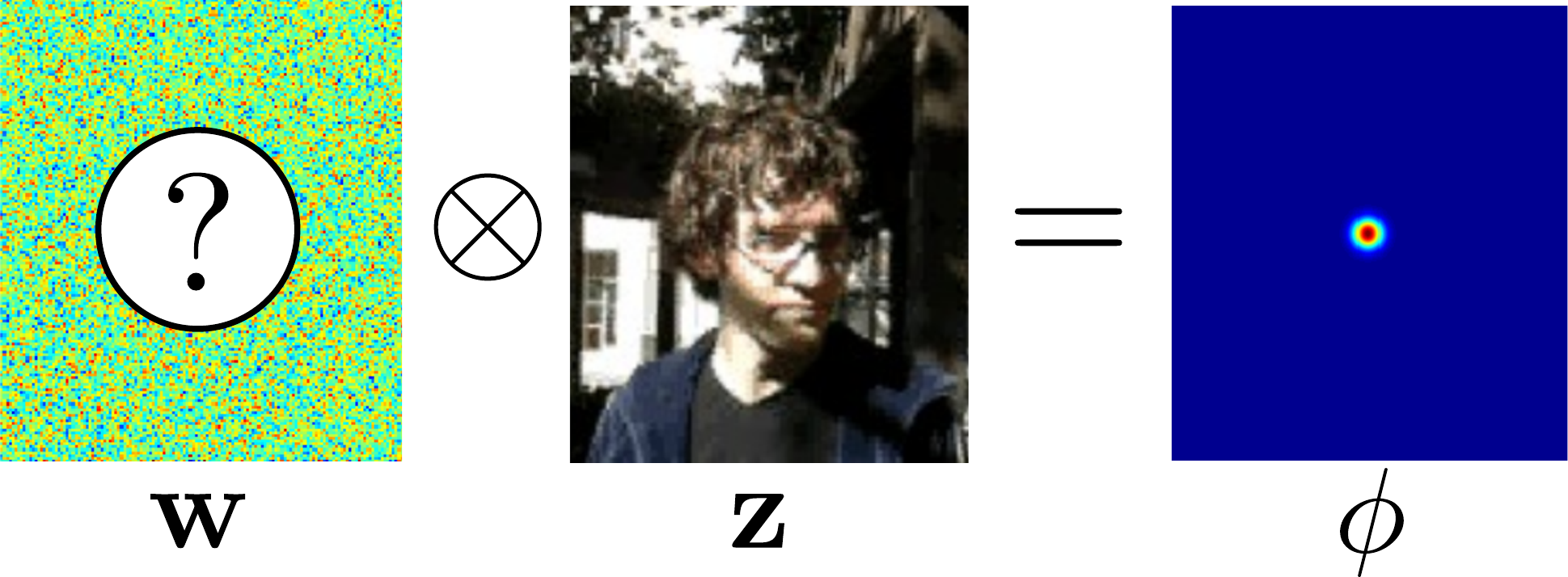}
\end{center}
   \caption{The correlation filter formulation. We seek a weight matrix $\mathbf{w}$ that results in a Gaussian response function $\phi$ when correlated over the image patch $\mathbf{z}$.}
\label{fig:kcf}
\end{figure}

This section summarizes the main results of the recent advances in correlation filters and their application to tracking~\cite{danelljan_color_cf_cvpr14,henriques2015tracking}. Given a single grayscale image patch $\mathbf{z}$ of size $M \times N$ a linear regression function $f(\mathbf{z})=\mathbf{w}^T \mathbf{z}$ is estimated such that its response is maximal at the center of the patch and gradually reduces for the patch circular shifts $\mathbf{z}_{m,n}$, $(m,n) \in \{0, \dots,M-1 \} \times \{0,\dots,N-1 \}$ toward the patch edge. This is formulated by minimizing the following cost function
\begin{equation}\label{eq:kcf_cost}
	\epsilon = || \mathbf{w} \otimes \mathbf{z} -\phi ||^2 + \lambda ||\mathbf{w}||^2,
\end{equation}
where $\otimes$ denotes circular correlation, $\phi$ is a Gaussian function centered at zero shift (see Figure~\ref{fig:kcf}) and $\lambda$ is a ridge regression regularization parameter which controls overfitting. 
The correlation in~(\ref{eq:kcf_cost}) is kernelized~\cite{henriques2015tracking} by redefining the $\mathbf{w}$ as a linear combination of the circular shifts, i.e., $\mathbf{w} = \sum\nolimits_{m,n} a_{m,n} \varphi(\mathbf{z}_{m,n})$, where $\varphi(\cdot)$ is a mapping to the Hilbert space induced by a kernel $\kappa(\cdot,\cdot)$. The minimum of (\ref{eq:kcf_cost}) is obtained at 
\begin{equation}\label{eq:kcf_solution}
	\boldsymbol{A} = \frac{ \boldsymbol{\Phi} }{ \boldsymbol{U}_z + \lambda },
\end{equation}
where the capital letters denote the Fourier transforms of image-domain variables, i.e., $\boldsymbol{A}=\mathcal{F}[\boldsymbol{a}]$, $\boldsymbol{\Phi}=\mathcal{F}[\boldsymbol{\phi}]$, $\boldsymbol{U}_z=\mathcal{F}[\boldsymbol{u}_z]$, with $\boldsymbol{u}_z(m,n)=\kappa( \mathbf{z}_{m,n}, \mathbf{z} )$ and $\boldsymbol{a}$ is a dual representation of $\mathbf{w}$~\cite{henriques2015tracking}. At time-step $t$, a patch $\mathbf{y}_t$ of size $M \times N$ is extracted from the image and the probability of object at pixel location $\mathbf{x}_t$ is calculated from the current estimate of $\boldsymbol{A}_t$ and the template $\mathbf{z}_t$ as 
\begin{equation}\label{eq:kcf_prob_response}
	p(\mathbf{y}_t|\mathbf{x}_t,\mathbf{z}_t) \propto \mathcal{F}^{-1}[ \boldsymbol{A}_t \odot \boldsymbol{U}_y],
\end{equation}
where $\boldsymbol{U}_y = \mathcal{F}[ \boldsymbol{u}_y ]$, $\boldsymbol{u}_y(m,n)=\kappa(\mathbf{y}_{m,n},\mathbf{z}_t)$. In \cite{danelljan_color_cf_cvpr14, henriques2015tracking}, the maximum on $p(\mathbf{y}_t|\mathbf{x}_t, \mathbf{z}_t)$ is taken as the new object position. The numerator and denominator of $\boldsymbol{A}_t$ in~(\ref{eq:kcf_solution}) as well as the patch template $\mathbf{z}_t$ are updated separately at the estimated position by an autoregressive model. The extension of the kernelized filter from grayscale patches to multi-channel features is straigth-forward and we refer the reader to~\cite{danelljan_color_cf_cvpr14,henriques2015tracking} for details.

\subsection{The coarse representation}\label{sec:coarseRepresent}
 
The coarse object representation in our appearance model consists of two high-level object models: the object global template $\mathbf{z}_t^{(0)}$ (a root correlation filter)   
and a global color model $C_t=\{ p(\mathbf{x}_t|f), p(\mathbf{x}_t|b)\}$, specified by the foreground and background color histograms, $p(\mathbf{x}_t|f)$ and $p(\mathbf{x}_t|b)$, respectively, where $\mathbf{x}_t$ denotes the pixel coordinates. These models are used in each tracking iteration to coarsely estimate the center $\mathbf{x}_t^{(0)}$ of the object bounding box within a specified search region (Figure~\ref{fig:overview}, step 1), which is subsequently refined by the mid-level representation (Section~\ref{sec:midLevelRepresent}). 
 
\begin{figure}[h!]
\begin{center}
\includegraphics[width=1\linewidth]{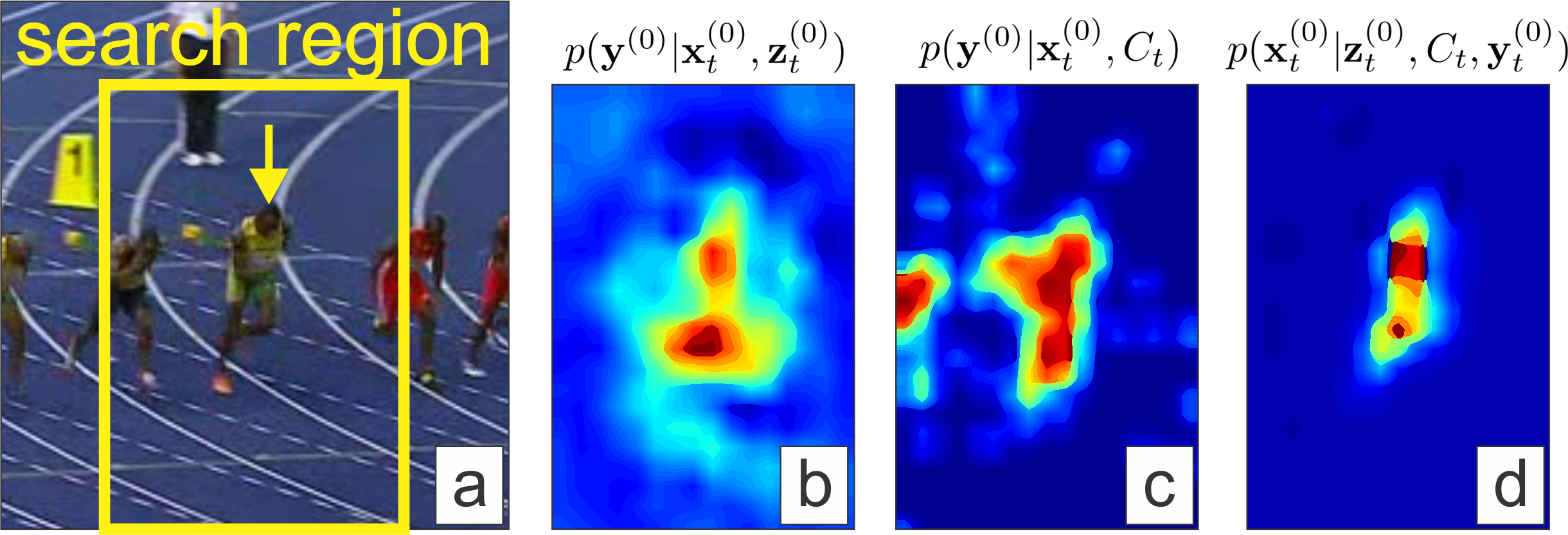}
\end{center}
   \caption{Example of a search region and the tracked object indicated by a rectangle and an arrow (a). The coarse template probability, the color probability and the full coarse model density are shown in (b), (c) and (d), respectively.}
\label{fig:segmentation_response}
\end{figure}

Given an image patch $\mathbf{y}_t^{(0)}$ extracted from a search region,   (Figure~\ref{fig:segmentation_response}a), the center is estimated by maximizing the probability of object location $\mathbf{x}_t^{(0)}$, 
\begin{equation}\label{eq:position_posterior_root}
	p(\mathbf{x}_t^{(0)}|\mathbf{z}_t^{(0)}, C_t, \mathbf{y}_t^{(0)} )\propto p(\mathbf{y}^{(0)}|\mathbf{x}_t^{(0)},\mathbf{z}_t^{(0)})p( \mathbf{y}^{(0)}|\mathbf{x}_t^{(0)}, C_t).
\end{equation}
The first term, $p(\mathbf{y}_t^{(0)}|\mathbf{x}_t^{(0)},\mathbf{z}_t^{(0)})$, is the template probability reflecting the similarity between the patch centered at $\mathbf{x}_t^{(0)}$ and the object template $\mathbf{z}_t^{(0)}$ calculated as the response from the correlation filter (\ref{eq:kcf_prob_response}), (see Figure~\ref{fig:segmentation_response}b). The second term is the color probability defined as 
\begin{equation}\label{eq:segmentation_uniform}
	p( \mathbf{y}^{(0)}|\mathbf{x}_t^{(0)}, C_t) = p(f|\mathbf{x}_t^{(0)}, \mathbf{y}_t^{(0)})(1-\alpha_\mathrm{col}) + \alpha_\mathrm{col},
\end{equation}
where $p(f|\mathbf{x}_t^{(0)}, \mathbf{y}_t^{(0)})$ is the probability of a pixel at location $\mathbf{x}_t^{(0)}$ belonging to a foreground and $\alpha_\mathrm{col}$ is a weak uniform distribution that addresses sudden changes of the object color, since the $p(f|\mathbf{x}_t^{(0)}, \mathbf{y}_t^{(0)})$ might be uninformative in these situations and would deteriorate localization. The value of $\alpha_\mathrm{col}$ varies with a color informativeness as detailed in Section~\ref{sec:color_informativeness}. The probability $p(f|\mathbf{x}_t^{(0)}, \mathbf{y}_t^{(0)})$ is calculated by histogram backprojection, i.e., by applying the Bayes rule with $p(\mathbf{x}_t|f)$ and $p(\mathbf{x}_t|b)$, and regularized by a Markov random field~\cite{kristan_segmentation_accv14, diplaros_genmodel} to arrive at a smoothed foreground posterior (Figure~\ref{fig:segmentation_response}c). Multiplying the template and color probabilities yields the density $p(\mathbf{x}_t^{(0)}|\mathbf{z}_t^{(0)}, C_t, \mathbf{y}_t^{(0)} )$ (Figure~\ref{fig:segmentation_response}d). Notice that on their own, the template and color result in ambiguous densities but their combination drastically reduces the ambiguity. 

\subsubsection{Color informativeness test}\label{sec:color_informativeness}

\begin{figure}[h!]
\begin{center}
\includegraphics[width=0.8\linewidth]{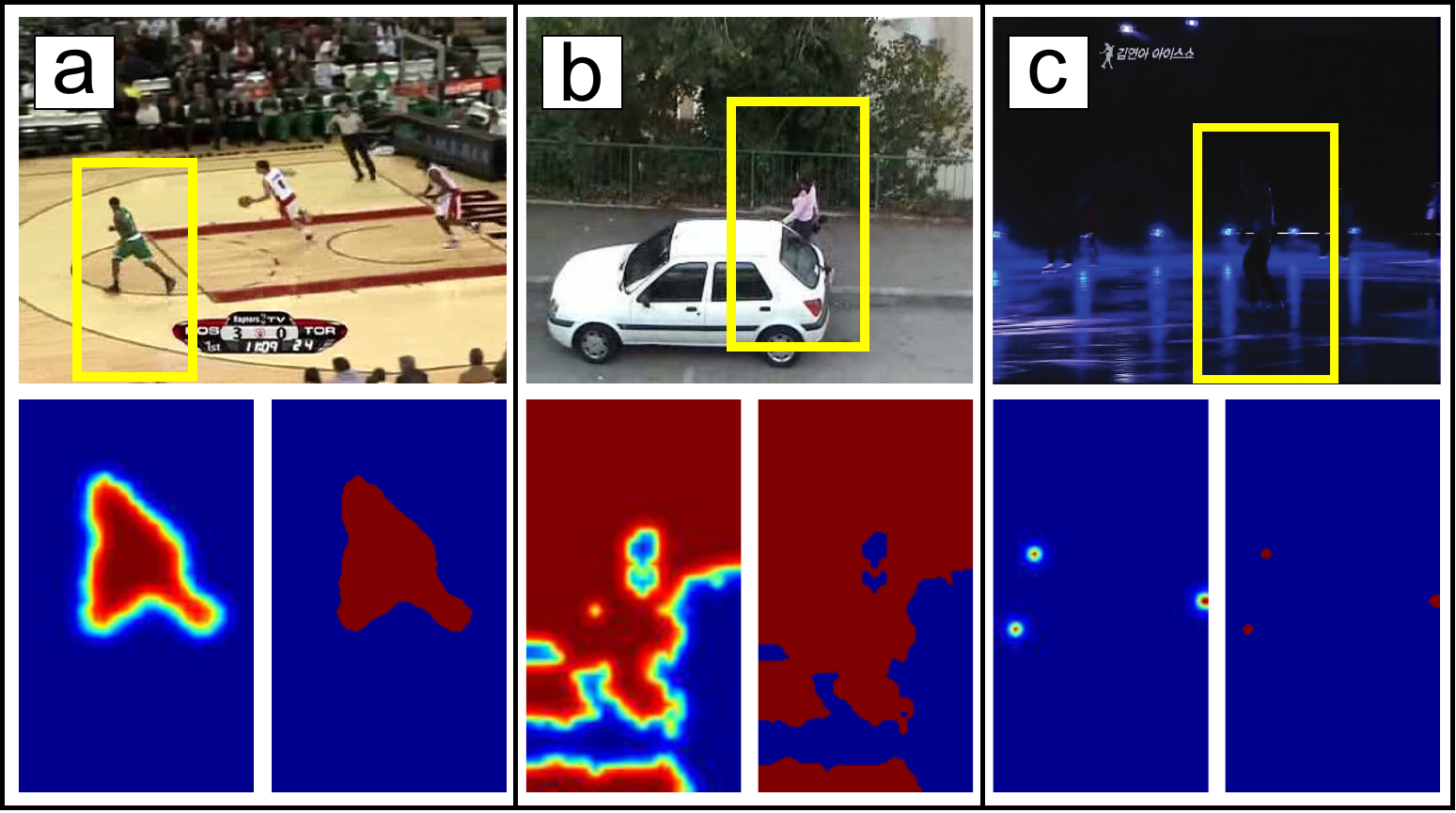}
\end{center}
   \caption{Three examples of the color backprojection within the image patch denoted with the yellow bounding box. The regularized backprojection is shown on left and the binarized segmentation on right under each image. Example (a) passes the color informativeness test, while (b) and (c) fail the test since too many or too few pixels are assigned to the object.}
\label{fig:segmentation_quality}
\end{figure}

Whenever the object color is similar to the background, or during sudden illumination variations, the color segmentation becomes unreliable and can degrade tracking performance. The color informativeness test is performed by comparing the number of pixels, $M^{(\mathrm{fg})}_t$, assigned to the foreground by the color model $p(f|\mathbf{x}_t^{(0)} \mathbf{y}_t^{(0)})$, and the object size from the previous time-step $M^{(\mathrm{siz})}_{t-1}$ (i.e., the area of object bounding box). If the deviation from the expected object area is within the allowed bounds, the uniform component in (\ref{eq:segmentation_uniform}) is set to a low value, otherwise it is set to 1, effectively ignoring the color information in the object position posterior (\ref{eq:position_posterior_root}), i.e.,
\begin{equation}\label{eq:uniform_component} 
    \alpha_\mathrm{col} = 
    \left\{
	\begin{array}{lll}
	0.1 &	; \alpha_\mathrm{min}  < \frac{M^{(\mathrm{fg})}_t}		{M^{(\mathrm{siz})}_{t-1}} < \alpha_\mathrm{max} \\
		1	&	; \mbox{otherwise}
	\end{array}
\right .
\end{equation}
The parameters $\alpha_\mathrm{min}$ and $\alpha_\mathrm{max}$ specify the  interval of expected number of pixels assigned to the target relative to the target bounding box size from the previous time-step. Since the aim of (\ref{eq:uniform_component}) is only to detect drastic segmentation failures, these values can be set to a very low and very large value, respectively. Figure~\ref{fig:segmentation_quality} illustrates the color informativeness test. In Figure~\ref{fig:segmentation_quality}(a), the number of pixels assigned to the foreground is within the expected bounds, while (b,c) show examples that fail the test by assigning too many or too few pixels to the object.
 
\subsection{The mid-level representation}\label{sec:midLevelRepresent}

The mid-level representation in our tracker is a geometrically constrained constellation of $N_p$ parts $\mathbf{X}_t=\{ \mathbf{x}_t^{(i)} \}_{i=1:N_p}$, where $\mathbf{x}_t^{(i)}$ is the position of $i$-th part (see Figure~\ref{fig:spring_system}, left). Note that the part sizes do not change during tracking and therefore do not enter the state variable $\mathbf{x}_t^{(i)}$.
Each part centered at $\mathbf{x}_t^{(i)}$ is a local mid-level representation of object, a kernelized correlation filter, specified by a fixed-size part template $\mathbf{z}_t^{(i)}$ and $\boldsymbol{A}_t^{(i)}$ (Section~\ref{sec:kcf}). 

The probability  of the constellation being at state $\mathbf{X}_t$ conditioned on the parts measurements \mbox{$\mathbf{Y}_t=\{ \mathbf{y}_t^{(i)} \}_{i=1:N_p}$} and parameters of the deformation model $\Theta$ is decomposed into
\begin{equation}\label{eq:posterior}
    p(\mathbf{X}_t|\mathbf{Y}_t,\Theta) \propto p(\mathbf{Y}_t|\mathbf{X}_t, \Theta)p(\mathbf{X}_t|\Theta).
\end{equation}
The density $p(\mathbf{Y}_t|\mathbf{X}_t, \Theta)$ is the \textit{measurement constraint} term, reflecting the agreement of measurements with the current state $\mathbf{X}_t$ of constellation, whereas the second term, $p(\mathbf{X}_t|\Theta)$, reflects the agreement of the constellation with the \textit{geometric constraints}.

\subsubsection{Geometric constraints}

The constellation is specified by a set of links $(i,j) \in \mathcal{L}$ indexing the connected pairs of parts (Figure~\ref{fig:spring_system}). The parts and links form an undirected graph and the joint pdf over the part states can be factored over the links as
\begin{equation}\label{eq:geomConstraints}
p(\mathbf{X}_t|\Theta)=\prod\nolimits_{(i,j)\in \mathcal{L}}\phi(||d_t^{(i,j)}|| ; \mu^{(i,j)}, k^{(i,j)}),
\end{equation}
where $d_t^{(i,j)}=\mathbf{x}_t^{(i)} - \mathbf{x}_t^{(j)}$ is a difference in positions of the linked parts, $\mu^{(i,j)}$ is the preferred distance between the pair of parts and $k^{(i,j)}$ is the intensity of this constraint. The factors in (\ref{eq:geomConstraints}) are defined as Gaussians $\phi(\cdot; \mu, k)$ with mean $\mu$ and variance $k$ meaning that deviations from the preferred distances decrease the probability (\ref{eq:geomConstraints}).

\subsubsection{Measurement constraints}

Given a fixed part state, $\mathbf{x}_t^{(i)}$, the measurement $\mathbf{y}_t^{(i)}$ at that part is independent from the states of other parts. The measurement probability decomposes into a product of per-part visual likelihoods
\begin{equation}\label{eq:measprob}
 p(\mathbf{Y}_t|\mathbf{X}_t, \Theta) = \prod\nolimits_{i=1:N_p} p(\mathbf{y}_t^{(i)} |\mathbf{x}_t^{(i)},\Theta).
\end{equation}
To simplify the combination of the geometric and the visual constraints (Section~\ref{sec:settingSprings}) it is beneficial to chose the visual likelihoods from the same class of functions as (\ref{eq:geomConstraints}). We make use of the fact that the parts appearance models are correlation filters trained on Gaussian outputs, thus the visual likelihoods in (\ref{eq:measprob}) can be defined as Gaussians as well. Let $\mathbf{x}_{tA}^{(i)}$ be the position in vicinity of $\mathbf{x}_t^{(i)}$ that maximizes the similarity of the appearance model $\mathbf{z}_t^{(i)}$ and the measurement $\mathbf{y}_t^{(i)}$ (see Figure~\ref{fig:spring_system}, left). The visual likelihood can then be defined as a Gaussian $p(\mathbf{y}^{(i)} |\mathbf{x}^{(i)},\Theta)=\phi(||d_t^{(i)}||; 0, k^{(i)})$ where $d_t^{(i)}=\mathbf{x}_{t}^{(i)}- \mathbf{x}_{tA}^{(i)}$ is the difference of the part current state and its visually-ideal position, and $k^{(i)}$ is the intensity of this constraint. 
 
\subsubsection{The dual spring-system formulation}\label{sec:settingSprings}

Substituting equations (\ref{eq:geomConstraints},\ref{eq:measprob}) back into (\ref{eq:posterior}) leads to an exponential posterior $p(\mathbf{X}_t|\mathbf{Y}_t,\Theta)\propto \exp(-E)$, with
\begin{equation}\label{eq:springsystemE}
E = \frac{1}{2}\sum\limits_{i=1:N_p} {k_{t}^{(i)}{{\left\| {d_{t}^{(i)}} \right\|}^2}}  + {\sum\limits_{i,j \in \mathcal{L}} {k_t^{(i,j)}(\mu _t^{(i,j)} - \left\| {d_t^{(i,j)}} \right\|)} ^2}.
\end{equation}

\begin{figure}[h!]
\begin{center}
\includegraphics[width=1.0\linewidth]{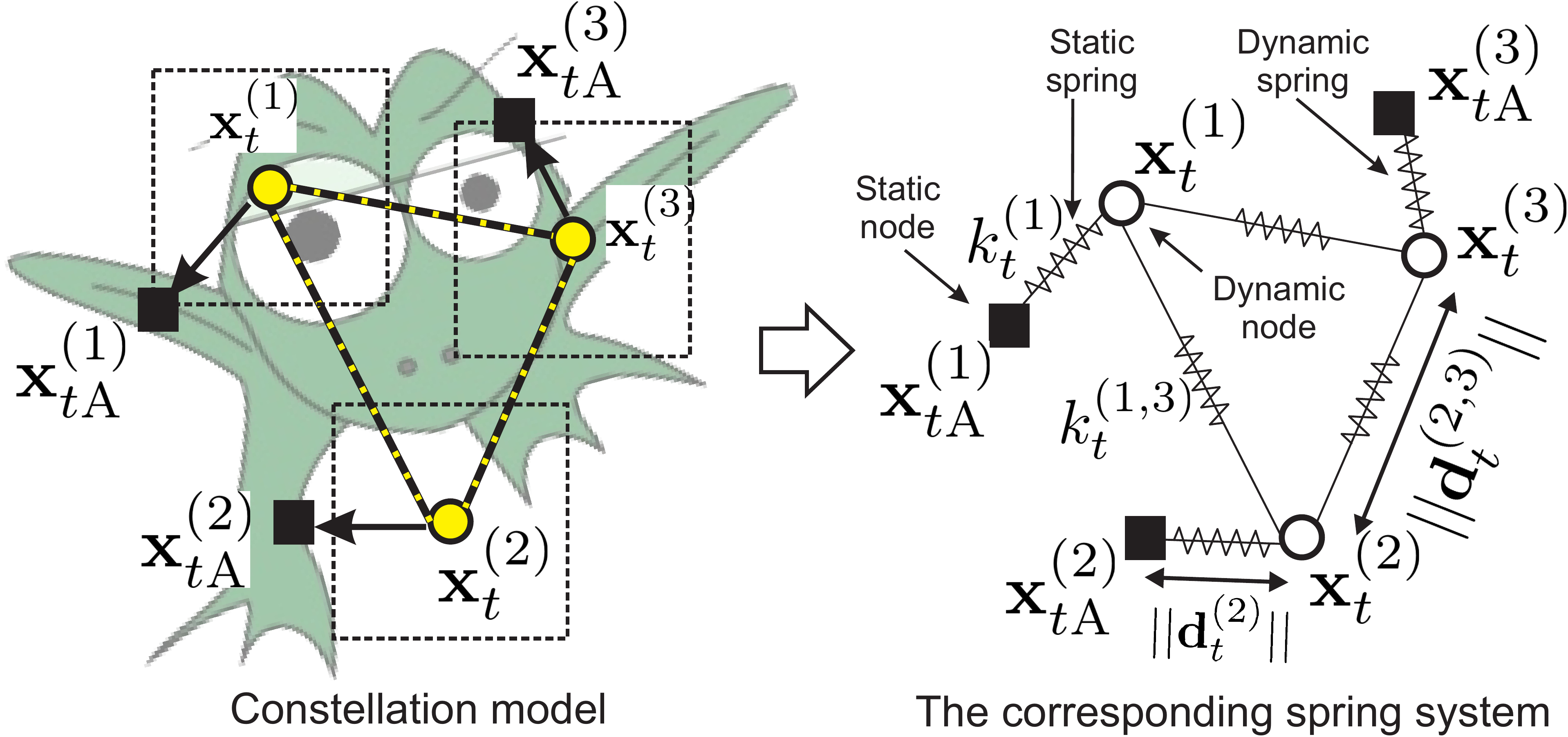}
\end{center}
   \caption{ Example of a constellation model with rectangular parts and arrows pointing to the most visually similar positions (left) and the dual form corresponding to a spring system (right). A constellation with only three nodes is shown for clarity.}
\label{fig:spring_system}
\end{figure}

Note that $E$ corresponds to an energy of a spring system in which pairs of parts are connected by springs and each part is connected by another spring to an image position most similar to the part appearance model (Figure~\ref{fig:spring_system}, right). The terms $\mu^{(i,j)}$ and $k^{(i,j)}$ are nominal lengths and stiffness of springs interconnecting parts (dynamic springs), while $k^{(i)}$ is stiffness of the spring connecting part to the image location (static spring). In the following we will refer to the nodes in the spring system that correspond to parts that move during optimization as \textit{dynamic nodes} and we will refer to the nodes that are anchored to image positions as \textit{static nodes}, since they do not move during the optimization.

The stiffness $k_t^{(i)}$ of a spring connecting a part to the image (in Figure~\ref{fig:spring_system} denoted as \emph{static spring}) should reflect the uncertainty of the visually best-matching location $\mathbf{x}_{tA}^{(i)}$ in the search region of the $i$-th part and is set by the output of the correlation filter. The best matching position $\mathbf{x}_{tA}^{(i)}$ is estimated as location at which the output of the corresponding correlation filter (\ref{eq:kcf_prob_response}) reaches a maximum value (denoted as $w_t^{(i)}$) and the spatial uncertainty in the search region is estimated as the weighted variance $\sigma_{t}^{2(i)}$, i.e., the average of squared distances from $\mathbf{x}_{tA}^{(i)}$ weighted by the correlation filter response map. The spring stiffness is thus defined by the response strength $w_t^{(i)}$ and spatial uncertainty, i.e.,
\begin{equation} \label{eq:stiffness_static}
	k_t^{(i)}=w_t^{(i)} / \sigma_{t}^{2(i)}.
\end{equation}

The stiffness of springs interconnecting the parts (in Figure~\ref{fig:spring_system} denoted as \emph{dynamic spring}) should counter significant deviations from the spring nominal length. Let $d_{tA}^{(i,j)} = \mathbf{x}_{tA}^{(i)}-\mathbf{x}_{tA}^{(j)}$ be the position difference between the visually most similar positions of the nodes indexed by $i$ and $j$. The stiffness of the spring connecting the nodes is set to
\begin{equation} \label{eq:stiffness_dynamic}
	k_t^{(i,j)} = \bigg(\frac{\mu_{t-1}^{(i,j)} - ||d_{tA}^{(i,j)}||}{ \mu_{t-1}^{(i,j)}}\bigg)^2.
\end{equation}
  
\subsection{Efficient MAP inference} \label{sec:springsOptimization}
 
\begin{figure}[h]
\begin{center}
\includegraphics[width=0.8\linewidth]{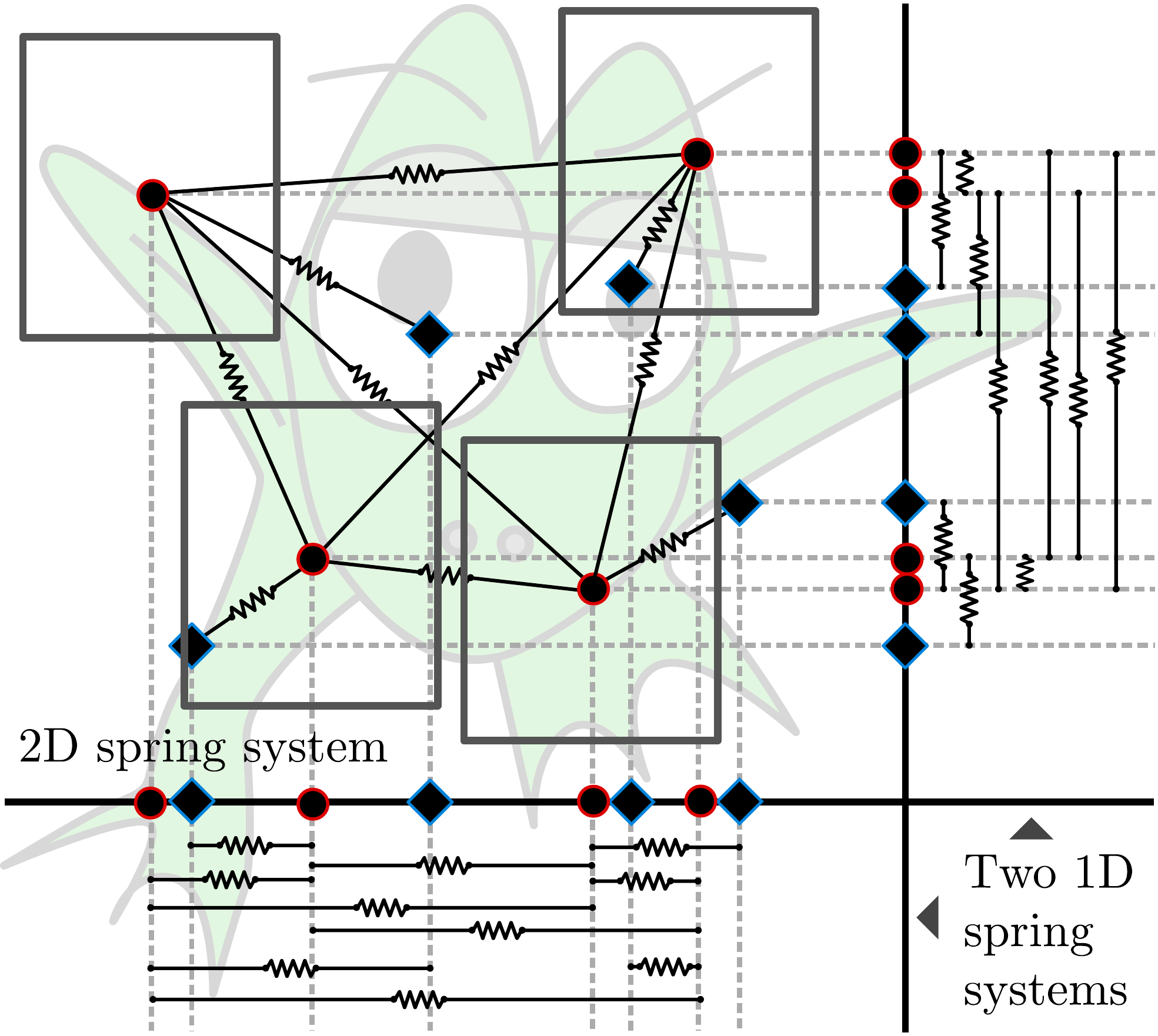}
\end{center}
   \caption{Example of decomposition of a 2D spring system with 4 dynamic nodes (circles) and 4 static nodes (diamonds) on two 1D spring systems. Each 1D spring system has a closed-form solution.}
\label{fig:spring_optimization}
\end{figure}

The spring system from Section~\ref{sec:settingSprings} is a dual representation of the deformable parts model and minimization of its (convex) energy function (\ref{eq:springsystemE}) corresponds to the maximum a posteriori state estimation (\ref{eq:posterior}) of the deformable parts model. This means that general-purpose convex energy minimizers can be used to infer the MAP state. But due to the dual spring system formulation, even more efficient optimizers can be derived. In particular, we propose an algorithm that splits a 2D spring system into two 1D systems, solves each in a closed form and then re-assembles them back into a 2D system (see Figure~\ref{fig:spring_optimization}). This partial minimization is iterated until convergence. In the following we derive an efficient closed-form solver for a 1D system.

Using standard results from Newtonian mechanics, the forces at springs $\mathbf{F}$ of a 1D spring system, can be written as
\begin{equation}\label{eq:spring_def1}
\textbf{F} = -\textbf{K}(\textbf{B}\textbf{x} - \textbf{L}),
\end{equation}
where $\textbf{K} = \mathrm{diag}([k_1,\cdots,k_N])$ is a diagonal matrix of spring stiffness coefficients, $\textbf{x}$ is a vector of 1D nodes positions, $\textbf{L} = [l_1, \cdots, l_N]$ is a vector of spring nominal lengths and $\mathbf{B}$ is a $N_{\mathrm{springs}}\times N_{\mathrm{nodes}}$ connectivity matrix that represents 
directed connections between the nodes. Let $\{n_{i1}, n_{i2}\}$ be indexes of two nodes connected by the $i$-th spring. The entries of $\mathbf{B}$ are then defined as  
\begin{equation}\label{eq:matrix_a_element}
b_{ij} = 
\left\{
	\begin{array}{lll}
		1	&	; j \equiv n_{i1} \\
		-1	&	; j \equiv n_{i2} \\
		0	&	; \mbox{otherwise}
	\end{array}
\right .
\end{equation}

The forces at nodes $\mathbf{F}_\mathrm{nodes}$ are given by left-multiplication of (\ref{eq:spring_def1}) by $\mathbf{B}^T$, yielding 
\begin{equation}\label{eq:spring_def3}
	\mathbf{F}_\mathrm{nodes} = -\textbf{B}^{T}\textbf{KBx} + \textbf{B}^{T}\textbf{KL}.
\end{equation}
The equilibrium is reached when the forces at nodes vanish (i.e., become zero), resulting in the following linear system 
\begin{equation}\label{eq:spring_linear_system}
	\widehat{\mathbf{K}} \mathbf{x} = \mathbf{C}\mathbf{L}, 
\end{equation}
where $\widehat{\mathbf{K}} = \mathbf{B}^{T}\mathbf{K}\mathbf{B}$ and $\mathbf{C} = \mathbf{B}^{T}\mathbf{K}$. We will assume the following ordering in the nodes positions vector, $\textbf{x} = [\textbf{x}_{\mathrm{dyn}}, \textbf{x}_{\mathrm{stat}}]^{T}$,
where $ \textbf{x}_{\mathrm{dyn}}$ and  $\textbf{x}_{\mathrm{stat}}$ are 1D positions of the dynamic and static nodes, respectively. The matrix $\widehat{\mathbf{K}}$  can be written as
\begin{equation}\label{eq:K_decompose}
	\widehat{\mathbf{K}} = \left[ \frac{ \widehat{\mathbf{K}}_\mathrm{dyn} ~ \widehat{\mathbf{K}}_\mathrm{stat} }{  \widehat{\mathbf{K}}_\mathrm{rem} } \right], 
\end{equation}
where $\widehat{\mathbf{K}}_\mathrm{dyn}$ and $\widehat{\mathbf{K}}_\mathrm{stat}$ are $N_\mathrm{dyn} \times N_\mathrm{dyn}$ and $N_\mathrm{dyn} \times N_\mathrm{stat}$ submatrices, respectively, realting the dynamic nodes to each other and the static nodes. Similar decomposition can be performed on $\mathbf{C}$,
\begin{equation}\label{eq:C_decompose}
	\mathbf{C} = \left[ \frac{ \mathbf{C}_\mathrm{dyn} }{  \mathbf{C}_\mathrm{stat} } \right].
\end{equation}
Substituting the definitions (\ref{eq:K_decompose}) and (\ref{eq:C_decompose}) into (\ref{eq:spring_linear_system}) yields the following closed form for the dynamic nodes positions $\mathbf{x}_\mathrm{dyn}$, 
\begin{equation}\label{eq:closed_form_x}
\textbf{x}_{\mathrm{dyn}} = \widehat{\textbf{K}}_{\mathrm{dyn}}^{-1} (\textbf{C}_{\mathrm{dyn}} \textbf{L} - \widehat{\textbf{K}}_{\mathrm{stat}} \textbf{x}_{\mathrm{stat}}).
\end{equation}

The optimization of a 2D spring system, which we call iterative direct approach (IDA), is summarized in the Algorithm~\ref{alg:optimization}. At each iteration, a 2D system is decomposed into separate 1D systems, each system is solved by~(\ref{eq:closed_form_x}) and the 2D system is re-assembled. The process is iterated until convergence. Note that $\widehat{\textbf{K}}_{\mathrm{stat}} \textbf{x}_{\mathrm{stat}}$ and $\widehat{\textbf{K}}_{\mathrm{dyn}}^{-1}$ can be calculated only once and remain unchanged during the optimization.

\begin{algorithm}[h!]
\begin{algorithmic}[1]
 \REQUIRE {~}\\
    Positions of dynamic and static nodes, $\textbf{x}_{\mathrm{dyn}}$ and $\textbf{x}_{\mathrm{stat}}$, stiffness vector $\mathbf{k}$ and adjacency matrix $\mathbf{B}$.
 \ENSURE {~}\\
    Equilibrium positions of dynamic nodes $\textbf{x}_{\mathrm{dyn}}$.
 \\\hspace{-0.6cm}\textbf{Procedure:}
 	\STATE For each dimension separately construct $\widehat{\textbf{K}}_{\mathrm{dyn}}$, 		
    	   $\widehat{\textbf{K}}_{\mathrm{stat}}$ and $\mathbf{C}_{\mathrm{dyn}}$ according to 			   (\ref{eq:K_decompose}) and (\ref{eq:C_decompose}).
	\WHILE{stop condition} 
		\STATE{For each dimension do:}
		\STATE{* Extract 1D positions of dynamic nodes from $\textbf{x}_{\mathrm{dyn}}$.}
		\STATE{* Calculate the current 1D spring lengths vector $\textbf{L}$.}
		\STATE{* Estimate new values of $\textbf{x}_{\mathrm{dyn}}$ by solving~(\ref{eq:closed_form_x}).}
        \STATE{Reassemble the 2D system.}
	\ENDWHILE
\end{algorithmic}
\caption{\label{alg:optimization}: Optimization of a 2D spring system.}
\end{algorithm}
 
\subsection{Deformable parts tracker (DPT)} \label{sec:tracking}

The coarse representation and the mid-level constellation of parts from Section~\ref{sec:coarseRepresent} and Section~\ref{sec:midLevelRepresent} are integrated into a tracker that localizes the object at each time-step within a search region by a top-down localization and bottom-up updates. In the following we will call this tracker a deformable parts correlation filter tracker and denote it by DPT for short. The tracker steps are visualized in Figure~\ref{fig:overview} and detailed in the following subsections.

\subsubsection{Top-down localization}\label{sec:td_localize}

The object is coarsely localized within a search region corresponding to the root correlation filter centered at the object position from the previous time-step $t-1$. The object center at time-step $t$ is approximated by position that maximizes the conditional probability $p(\mathbf{x}_t^{(0)}|\mathbf{z}_t^{(0)},C_t,\mathbf{y}_t^{(0)})$ from Section~\ref{sec:coarseRepresent} and a coarse center translation from $t-1$ to $t$ is estimated (Figure~\ref{fig:overview}, step 1). The mid-level representation, i.e, constellation of parts, is initialized by this translation. For each translated part $\mathbf{x}_t^{(i)}$, the part correlation filter is applied to determine the position of the maximum similarity response, $\mathbf{x}_{tA}^{(i)}$, along with the stiffness coefficients $k_{t}^{(i)}$ and $k_{t}^{(i,j)}$ as detailed in Section~\ref{sec:settingSprings}. A MAP constellation estimate $\hat{\mathbf{X}}_t$ is obtained by minimizing the energy (\ref{eq:springsystemE}) of the equivalent spring system optimization from Section~\ref{sec:springsOptimization} (Figure~\ref{fig:overview}, steps 2-4).

\subsubsection{Bottom-up update}\label{sec:bu_update}

The mid-level and coarse representations are updated as follows (Figure~\ref{fig:overview}, steps 5,6). The part correlation filters and their appearance models $\mathbf{z}_t^{(i)}$ are updated at MAP estimates of part positions $\hat{\mathbf{x}}_t^{(i)}$. Updating all appearance models at constant rate might lead to drifting and failure whenever the object is partially occluded or self-occluded. An effective mechanism is applied to address this issue. A part is updated only if its response at the MAP position $\hat{\mathbf{x}}_t^{(i)}$ is at least half of the strongest response among all parts and if at least twenty percent of all pixels within the part region correspond to the object according to the segmentation mask estimated at the root part (Section~\ref{sec:coarseRepresent}). The nominal spring lengths (the preferred distances between parts) are updated by an autoregressive scheme 
\begin{equation}
\mu_t^{(i,j)}= {\mu}_{t-1}^{(i,j)} (1-\alpha_\mathrm{spr}) + ||\hat{d}_{t}^{(i,j)}||\alpha_\mathrm{spr},
\end{equation}
where $||\hat{d}_{t}^{(i,j)}||$ is the distance between the parts $(i,j)$ in the MAP estimate $\hat{\mathbf{X}}_t$ and $\alpha_\mathrm{spr}$ is the update factor.

The coarse representation is updated next. The MAP object bounding box is estimated by $\hat{\mathbf{x}}_t^{(0)}=\mathbf{T}_t\hat{\mathbf{x}}_{t-1}^{(0)}$, where $\mathbf{T}_t$ is a Euclidean transform estimated by least squares from the constellation MAP estimates $\hat{\mathbf{X}}_{t-1}$ and $\hat{\mathbf{X}}_t$. The root correlation filter $\mathbf{z}_t^{(0)}$ and the histograms in the global color model $C_t$ are updated at $\hat{\mathbf{x}}_t^{(0)}$. A histogram $\mathbf{h}_t^{(f)}$ is extracted from $\hat{\mathbf{x}}_t^{(0)}$ and another histogram $\mathbf{h}_t^{(b)}$ is extracted from the search region surrounding $\hat{\mathbf{x}}_t^{(0)}$ increased by a factor $\alpha_{\mathrm{sur}}$. The foreground and background histograms are updated by an autoregressive model, i.e.,
\begin{equation}\label{eq:autoregressive_update}
	p(\mathbf{x}_t| \cdot)=p(\mathbf{x}_{t-1}|\cdot) (1-\alpha_\mathrm{hist}) +\mathbf{h}_t^{(\cdot)}\alpha_\mathrm{hist}, 
\end{equation}
where $\alpha_\mathrm{hist}$ is the forgetting factor. To increase adaptation robustness, the histograms are not updated if the color segmentation fails the color informativeness test from Section~\ref{sec:color_informativeness}. The top-down localization and bottom-up update steps are summarized in Algorithm~\ref{alg:process}.

\begin{algorithm}[h!]
\begin{algorithmic}[1]
 \REQUIRE {~}\\
    Coarse model $\{  \mathbf{x}^{(0)}_{t-1}, \mathbf{z}^{(0)}_{t-1}, \mathbf{C}_{t-1}\}$ and mid-level model $\{ \mathbf{X}_{t-1}, \mathbf{Z}_{t-1}\}$ at time-step $t-1$.
 \ENSURE {~}\\
    Coarse model $\{  \mathbf{x}^{(0)}_{t}, \mathbf{z}^{(0)}_{t}, \mathbf{C}_{t}\}$ and mid-level model $\{ \mathbf{X}_{t}, \mathbf{Z}_{t}\}$ at time-step $t$.
 \\\hspace{-0.6cm}\textbf{Procedure:}
    \STATE Coarsely estimate the object position by the root node 
    	   (Section~\ref{sec:coarseRepresent}) and displace the mid-level parts.       
    \STATE Calculate the part correlation filter responses and form a spring system according 
    	   to Section~\ref{sec:settingSprings}. 
    \STATE Estimate the MAP mid-level parts constellation by optimizing the energy of a dual 			spring system (Section~\ref{sec:springsOptimization}).
    \STATE Update the root node position and size by the Euclidean transform fitted to the parts positions before and after MAP inference 				 	 
    		(Section~\ref{sec:td_localize}).    
    \STATE Update the spring system parameters and the constellation appearance models 
    		(Section~\ref{sec:bu_update}).
    \STATE Update the coarse color model $\mathbf{C}_t$ and correlation filter 
    		$\mathbf{z}_t^{(0)}$.
\end{algorithmic}
\caption{\label{alg:process}: A tracking iteration of a deformable parts correlation filter tracker.}
\end{algorithm}

\subsubsection{Tracker initialization} \label{sec:initialization_details}

The coarse representation at time-step $t=1$ is initialized from the initial bounding box $\mathbf{x}_1^{(0)}$. The mid-level the constellation of parts is initialized by splitting the initial object bounding box into four equal non-overlapping parts. The part appearance models are initialized at these locations and the preferred distances between parts are calculated from the initialized positions. 
 
\section{Experimental analysis}

This section reports experimental analysis of the proposed DPT. The implementation details are given in Section~\ref{sec:implementation}, Section~\ref{sec:ldp_analysis} details the analysis of the design choices, Section~\ref{sec:sota} reports comparison to the related state-of-the-art, Section~\ref{sec:benchmarks} reports performance on recent benchmarks and Section~\ref{sec:qualitative} provides qualitative analysis.
 
\subsection{Implementation details and parameters}\label{sec:implementation}
 
Our implementation uses a kernelized correlation filters (KCF)~\cite{henriques2015tracking} with HOG~\cite{dalal_triggs_hog} features and grayscale template in the part appearance models. All filter parameters and learning rate are the same as in~\cite{henriques2015tracking}. The parts have to be large enough to capture locally visually-distinctive regions on the object and have to cover the object without significantly overlapping with each other. The size of the tracked targets therefore places a constraint on the maximal number of parts since their size reduces with this number. For small parts, the HoG features become unreliable.
But even more pressing is the issue that the capture range of correlation filters is constrained by the template size and is even reduced in practice due to the effects of circular correlation used for learning and matching. Therefore, small parts increasingly lose the ability to detect large displacements. The parts have to be large enough to capture the object partial appearance at sufficient level of detail, therefore we set the number of parts to $N_p=4$. The DPT allows any type of connectivity among the parts and our implementation applies a fully-connected constellation for maximally constrained geometry. The foreground/background models $C_t$ are HSV color histograms with $16\times16\times16$ bins. The remaining parameters are as follows:
the rate of spring system update is $\alpha_\mathrm{spr}=0.95$,
the background histogram extraction area parameter is set to $\alpha_\mathrm{sur}=1.6$ and
the histogram update rate is set to $\alpha_\mathrm{hist}=0.05$.
These parameters have a straight-forward interpretation, were set to the values commonly used in published related trackers. Recall that the color informativeness test from Section~\ref{sec:color_informativeness} detects drastic segmentation failures. In our implementation the failure is detected if the number of pixels pixels assigned to the object relative to the target bounding box size either falls below 20 percent or exceeds the initial size by 100 percent, i.e.,  $\alpha_\mathrm{min}=0.2$ and $\alpha_\mathrm{max}=2.0$. Note that these are very weak constraints meant to detect obvious segmentation failures and did not require special tuning. The parameters have been fixed throughout all experiments.


The DPT was implemented in Matlab with backprojection and HoG extraction implemented in C and performed at $19$ FPS on an Intel Core i7 machine. Since our tracker uses a KCF~\cite{henriques2015tracking} for root and part appearance models, the complexity of our tracker is in order of the KCF complexity, which is $\mathcal{O}(n\log n)$, where $n$ is the number of pixels in the search region. The DPT has complexity five times the KCF, because of the four mid-level parts plus a root part. The localization and update of five KCFs takes approximately $40ms$. Our tracker consists also of the spring system and object segmentation. The optimization of the spring system takes on average less than $3ms$ and the color segmentation with the histogram extraction requires approximately $9ms$.

\subsection{The DPT design analysis} \label{sec:ldp_analysis}

 
\subsubsection{Analysis of the spring system optimization} \label{sec:spring_optimization_comparison}

This section analyzes the iterated direct approach (IDA) from Section~\ref{sec:springsOptimization}, which is the core of our part-based optimization. The following random spring system was used in the experiments. Dynamic nodes were initialized at uniformly distributed positions in a 2D region  $[0,1] \times [0,1]$. Each node was displaced by a randomly sampled vector $\mathbf{d} = [d_x, d_y] \sim \mathcal{U}([-0.5; 0.5])$ and the anchor nodes were set by displacing the corresponding dynamic nodes by the vector $\mathbf{b} = [b_x, b_y]\sim \mathcal{U}([-0.25; 0.25])$. The stiffness of $i$-th dynamic spring was set to $k_i = (\sigma d_i)^{-2}$, where $d_i$ is the length of the spring and $\sigma=0.1$ is the size change. The stiffness of $j$-th static spring was set to $k_j = \frac{1}{2} + u_j \overline{k}_{\mathrm{dyn}}$, where $\overline{k}_{\mathrm{dyn}}$ is the average stiffness of the dynamic springs and $u_j \sim \mathcal{U}([0; 1])$. The IDA was compared with the widely used conjugate gradient descent optimization (CDG), which guarantees a global minimum will be reached on a convex cost function and has shown excellent performance in practice on non-convex functions as well~\cite{linear_nonlinear_optimization}. All results here are obtained by averaging the performance on 100,000 randomly generated spring systems. 

\begin{figure}[h]
\begin{center}
\includegraphics[width=1\linewidth]{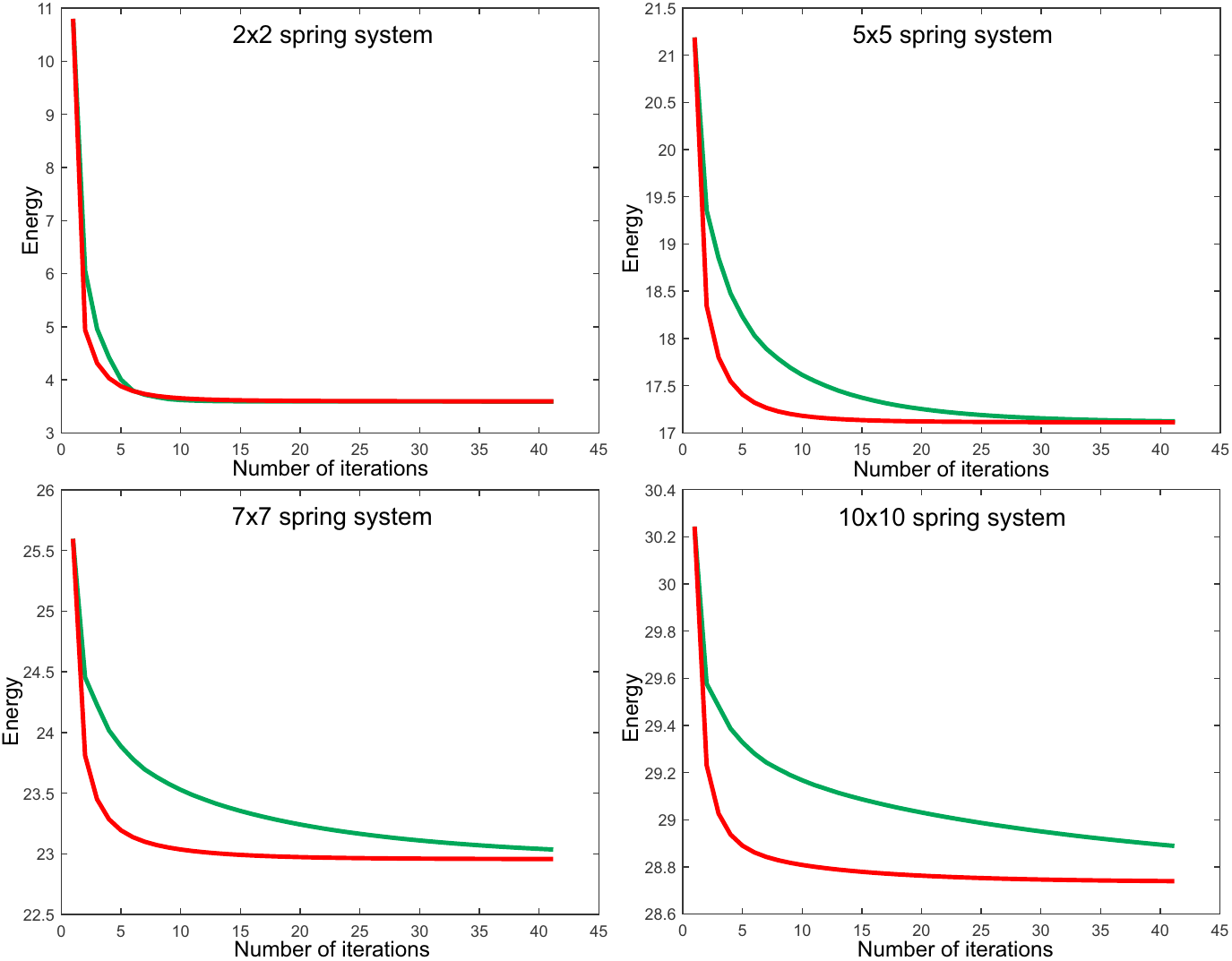}
\end{center}
   \caption{The spring system remaining energy w.r.t. the iterations. Experiment is averaged over 10,000 random spring systems. The red and the green curves represent IDA and CGD methods, respectively.}
\label{fig:cgd_ida_energy}
\end{figure}

The first experiment evaluated the convergence properties of IDA. Figure~\ref{fig:cgd_ida_energy} shows the energy reduction in spring system during optimization for different number of nodes in the spring system. The difference in the remaining energy after many iterations is negligible between CGD and IDA, which means that both converged to equivalent solutions. But the difference in energy reduction in consecutive steps and the difference in steps required to reach convergence is significant. The IDA reduces the energy at much faster rate than CGD and this result is consistent over various spring system sizes. Notice that IDA significantly reduced the energy already within the first few iterations. 

\begin{figure}[h!]
\begin{center}
\includegraphics[width=1\linewidth]{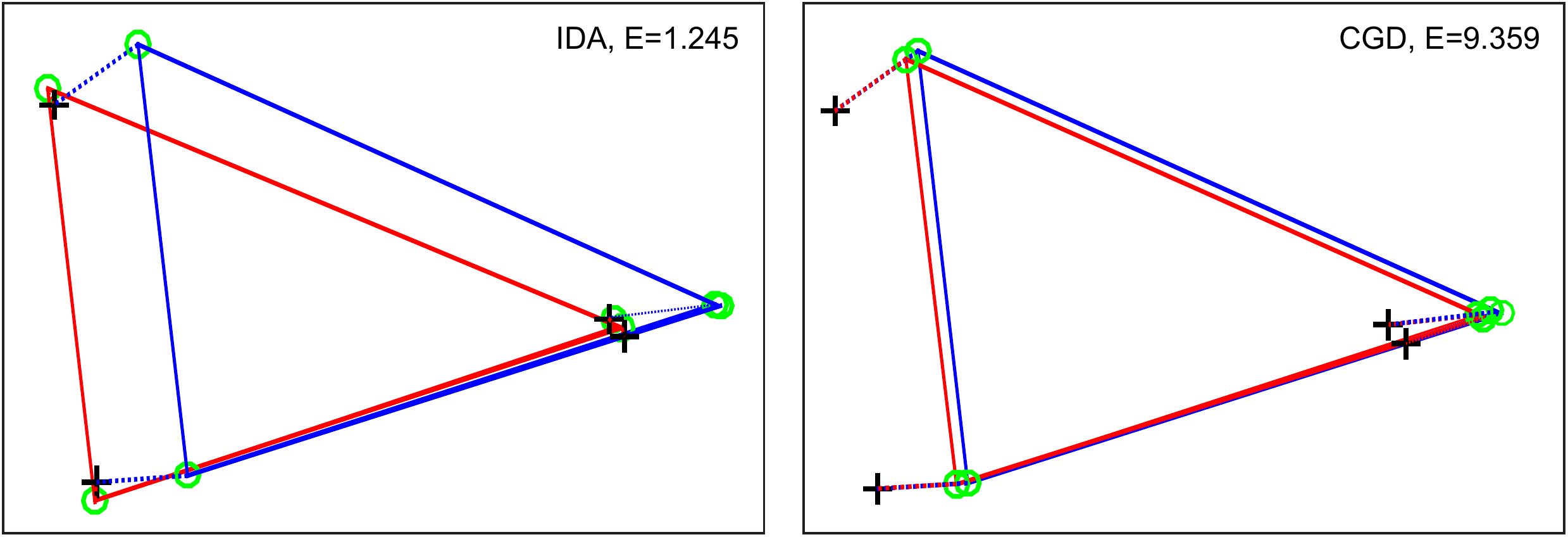}
\end{center}
   \caption{The dynamic part of the spring system before and after optimization is shown in blue and red, respectively. Dynamic nodes and anchor nodes are depicted by green circles and black crosses, respectively, and the black dotted lines depict the static springs. The remaining energy $E$ of the optimized spring system is shown as well.}
\label{fig:springs-optimization}
\end{figure}

The numeric behavior of IDA is much more robust than that of the CGD. Figure~\ref{fig:springs-optimization} shows an example of a spring system, where CGD did not reach the optimal state, but the IDA converged to a stable state with much lower energy, than the CGD. The poor convergence in CGD is caused by the very small distance between a pair of nodes compared to the other distances resulting in poor gradient estimation, while the IDA avoids this by the closed-form solutions for the marginal 1D spring systems. The IDA converged in 5 iterations, while the CGD stopped after 471 iterations. The spring systems like the one described here were automatically detected and removed in the simulated experiment to prevent skewing results for the CGD. The results conclusively show that the IDA converges to a global faster than CGD and is more robust.
 
\begin{figure}[h]
\begin{center}
\includegraphics[width=1\linewidth]{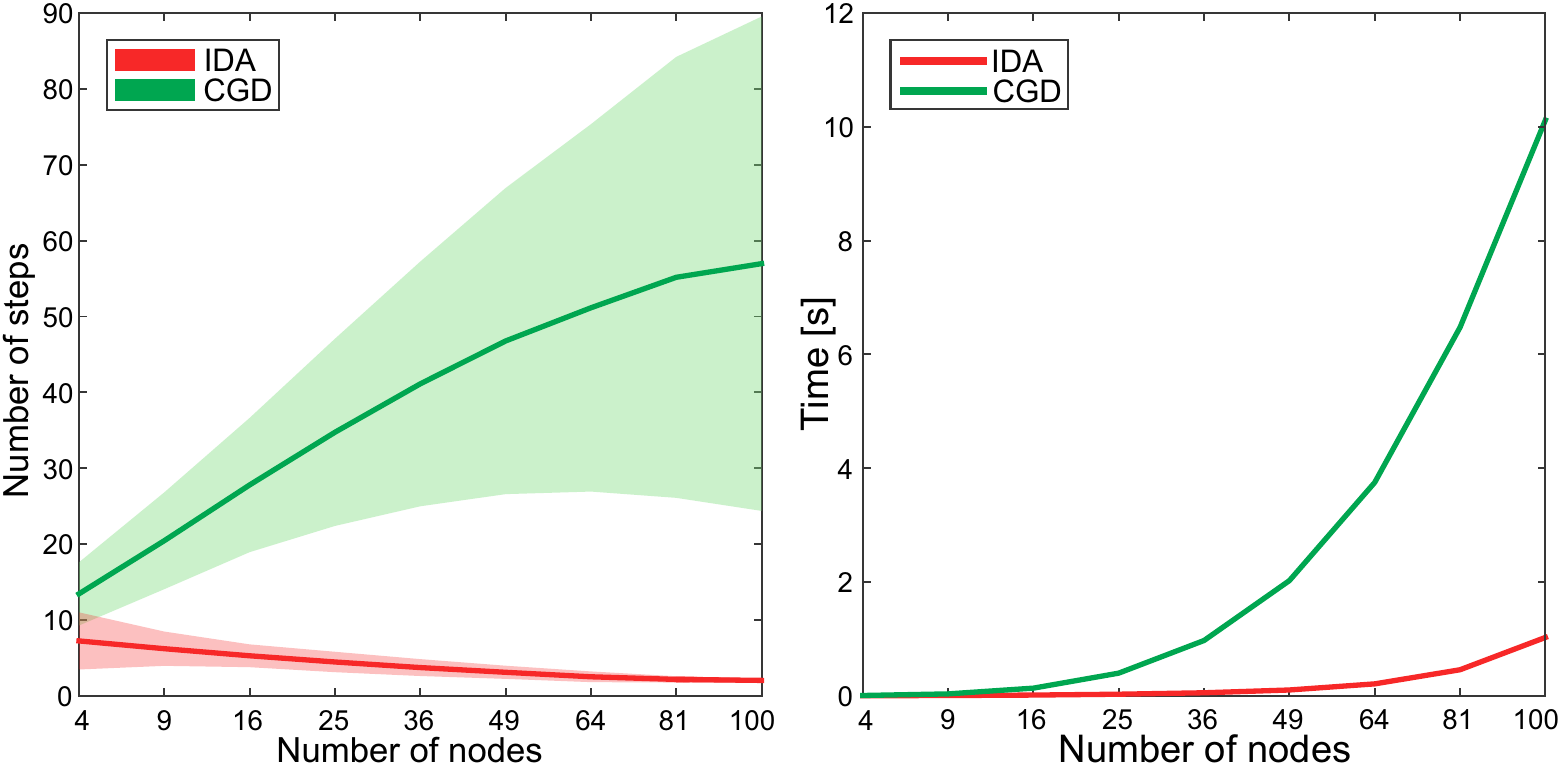}
\end{center}
    \caption{The number of iterations (left) and time (right) spent by IDA and CGD on optimization with respect to the spring system size.}
\label{fig:cgd_ida_complexity}
\end{figure}

The second experiment evaluated the IDA scalability. Figure~\ref{fig:cgd_ida_complexity} shows the optimization speed w.r.t. the spring system size. The number of iterations significantly increases for the CGD with increasing the number of parts. On the other hand, the IDA exhibits remarkable scalability by keeping the number of steps approximately constant over a range of system sizes. Furthermore, the variance in the number of iterations is kept low and consistently much lower than for the CGD. The iteration step complexity is expected to increase with the number of parts, since larger systems are solved. Figure~\ref{fig:cgd_ida_complexity} also shows that the computation times indeed increase exponentially for CGD, but the IDA hardly exhibits increase for a range of spring system sizes. These results conclusively show that IDA scales remarkably well. 

\subsubsection{The DPT parameters analysis} \label{sec:ldp_variants}
 
The DPT design choices were evaluated on a state-of-the-art short-term tracking benchmark VOT2014~\cite{kristan_vot2014,Kristan2015}. In contrast to related benchmarks that aim at large datasets, the datasets in VOT initiative~\cite{Kristan2015} are constructed by focusing on the challenging, well annotated, sequences while keeping the dataset small. The objects are annotated by rotated bounding boxes and all sequences are per-frame annotated by visual attributes. The VOT evaluation protocol initializes the tracker from a ground truth bounding box. Once the overlap between the ground truth and tracker output bounding box falls to zero, a failure is detected and tracker is re-initialized. The VOT toolkit measures two basic tracking performance aspects: reset-based accuracy and robustness. The reset-based accuracy is measured as the average overlap during successful tracking, while the robustness measures the number of failures (i.e., number of tacker re-sets). Apart from reporting raw accuracy/robustness values, the benchmark can rank trackers with respect to these measures separately by taking into account the statistical as well as practical difference. Since 2015 the VOT primary overall accuracy measure is the expected average overlap (EAO). This measure calculates the expected overlap on fixed-length sequences that a tracker would attain without reset. In addition we also report the primary OTB~\cite{otb_cvpr2010} measure. The OTB performance evaluation primarily differs from the VOT~\cite{kristan_vot2015} in that trackers are not reset at failure. The overall performance is reported by an average overlap (AO) over all sequences.


The first experiment analyzed the contributions of the proposed segmentation in the coarse layer and the lower-layer constellation model. The baseline tracker was a DPT variant that does not use the constellation, nor the segmentation ($\mathrm{DPT}_\mathrm{crs}^\mathrm{nos}$), which is in fact the original KCF~\cite{henriques2015tracking} correlation filter. Adding a segmentation model to the baseline tracker results in the coarse layer in our part-based tracker, which we denote by $\mathrm{DPT}_\mathrm{crs}$. Table~\ref{tbl:LDPvariants} clearly shows that the number of failures is reduced by our segmentation and the overall accuracy (EAO and AO) increases for $\mathrm{DPT}_\mathrm{crs}^\mathrm{nos}$. By adding the lower layer to the $\mathrm{DPT}_\mathrm{crs}$, we arrive at the proposed DPT, which further boosts the performance by all measures. In particular, the number of failures is reduced by over $4\%$, the reset-based accuracy increases by over $10\%$, the expected average overlap (EAO) increases by $8\%$ and the OTB average overlap (AO) increases by $10\%$. The VOT ranking methodology was applied to these three trackers. The DPT was ranked as the top-performing tracker, which conclusively shows that the improvements are statistically as well as practically significant. 
\begin{table}[h!]
\caption{\label{tbl:LDPvariants} Performance of DPT variants in terms of
raw reset-based accuracy (res. acc.) and robustness (rob.), the VOT rank, the VOT no-reset accuracy (expected average overlap, EAO) and the OTB no-reset average overlap (AO). The arrows $\uparrow$ and $\downarrow$ indicate that ``higher is better" and ``lower is better", respectively.}
\begin{center}
\begin{tabular}{l | c | c c | c | c} 
DPT & VOT &  \multicolumn{2}{c|}{Raw values}  &  {VOT } & OTB \\
variant & EAO$\uparrow$  &  res. acc.$\uparrow$  & rob. $\downarrow$  &  rank $\downarrow$ & AO $\uparrow$  \\ \hline 
\rowcolor{LightGray}
 DPT                        				& 0.39 & 0.61 & 0.47 & 1.42 & 0.486  \\
\rowcolor{DarkGray}
$\mathrm{DPT}_\mathrm{crs}$ 				& 0.36 & 0.55 & 0.49 & 2.10 & 0.442 \\
\rowcolor{DarkGray}
$\mathrm{DPT}_\mathrm{crs}^\mathrm{nos}$    & 0.21 &  0.57 & 1.13 & 3.06 & 0.377 \\ 
\rowcolor{LightGray}
$\mathrm{DPT}_\mathrm{str}$ 				& 0.34 & 0.57 & 0.61 &  2.06 & 0.467 \\
\rowcolor{LightGray}
$\mathrm{DPT}_\mathrm{loc}$ 				& 0.36 & 0.62 & 0.65 & 1.46  & 0.485 \\ 
\rowcolor{DarkGray}
$\mathrm{DPT}_{3\times 3}^\mathrm{ov}$      & 0.31 & 0.60 & 0.71 & 1.82 & 0.481 \\
\rowcolor{DarkGray}
$\mathrm{DPT}_{3\times 3}^\mathrm{nov}$	    & 0.31 & 0.60 & 0.73 & 1.90 & 0.481 \\
\end{tabular}
\end{center}
\end{table}

The DPT variants with fully connected, locally connected and star-based topology, $\mathrm{DPT}$, $\mathrm{DPT}_\mathrm{loc}$, $\mathrm{DPT}_\mathrm{str}$, respectively, were compared to evaluate the influence of the lower-layer topology. The
top performance in terms of the VOT EAO as well as OTB AO is achieved by the fully-connected topology, followed by the locally-connected and star-based topology. This order remains the same under the VOT ranking methodology, which confirms that the improvements of the fully-connected topology over the alternatives are statistically as well as practically significant.

For completeness, we have further tested the DPT performance with the increased number of parts at the lower layer. Given the constraints imposed on the parts size (as discussed in Section~\ref{sec:implementation}), we tested two variants with $3\times3=9$ parts: one with overlapping parts of the same size as in the original DPT ($\mathrm{DPT}_{3\times 3}^\mathrm{ov}$) and one with smaller, non-overlapping, parts ($\mathrm{DPT}_{3\times 3}^\mathrm{nov}$). Table~\ref{tbl:LDPvariants} shows that these versions of DPT perform similarly in terms of overall performance (EAO and AO), with $\mathrm{DPT}_{3\times 3}$ obtaining slightly better rank, which is due to slightly better robustness than $\mathrm{DPT}_{3\times 3}^\mathrm{ov}$. Both variants are outperformed by the original $2\times 2$ DPT. The improvement of DPT over the best $\mathrm{DPT}_{3\times 3}$ tracker is over $20\%$ in terms of the expected average overlap and approximately $2\%$ in terms of the OTB average overlap. The smaller difference in OTB AO is because $\mathrm{DPT}_{3\times 3}$ has a similar accuracy as 
$\mathrm{DPT}$, but fails more often. The OTB AO effectively measures the accuracy only up to the first failure. But the raw values clearly show superior robustness in $\mathrm{DPT}$ which is reflected in EAO. 
 
\subsection{Comparison to the state-of-the-art baselines} \label{sec:sota}

The DPT tracker is a layered deformable parts correlation filter, therefore we compared it to the state-of-the-art part-based as well as holistic discriminative trackers. The set of baselines included: (i) the recent state-of-the-art part-based baselines, PT~\cite{ruiyao_partbased_cvpr2013}, DGT~\cite{dgt2014tip}, CMT~\cite{cmt_nabehay_cvpr2015} and LGT~\cite{lgt_tpami2013}, (ii) the state-of-the-art discriminative baselines TGPR~\cite{gaoECCV2014}, Struck~\cite{hare_struck}, DSST~\cite{danelljan2014accurate}, KCF~\cite{henriques2015tracking} SAMF~\cite{samf_eccv2014}, STC~\cite{zhang_stc_eccv2014}, MEEM~\cite{meem_eccv14}, MUSTER~\cite{muster_cvpr15} and HRP~\cite{hrp_iccv15}, and (iii) the standard baselines CT~\cite{zhang_ct}, IVT~\cite{ross_ivt}, MIL~\cite{babenko_mil}. This is a highly challenging set of recent state-of-the-art containing all published top-performing trackers on VOT2014, including the winner of the challenge DSST~\cite{danelljan2014accurate} and trackers recently published at major computer vision conferences and journals.

\begin{figure}[ht!]
\begin{center}
\includegraphics[width=\linewidth]{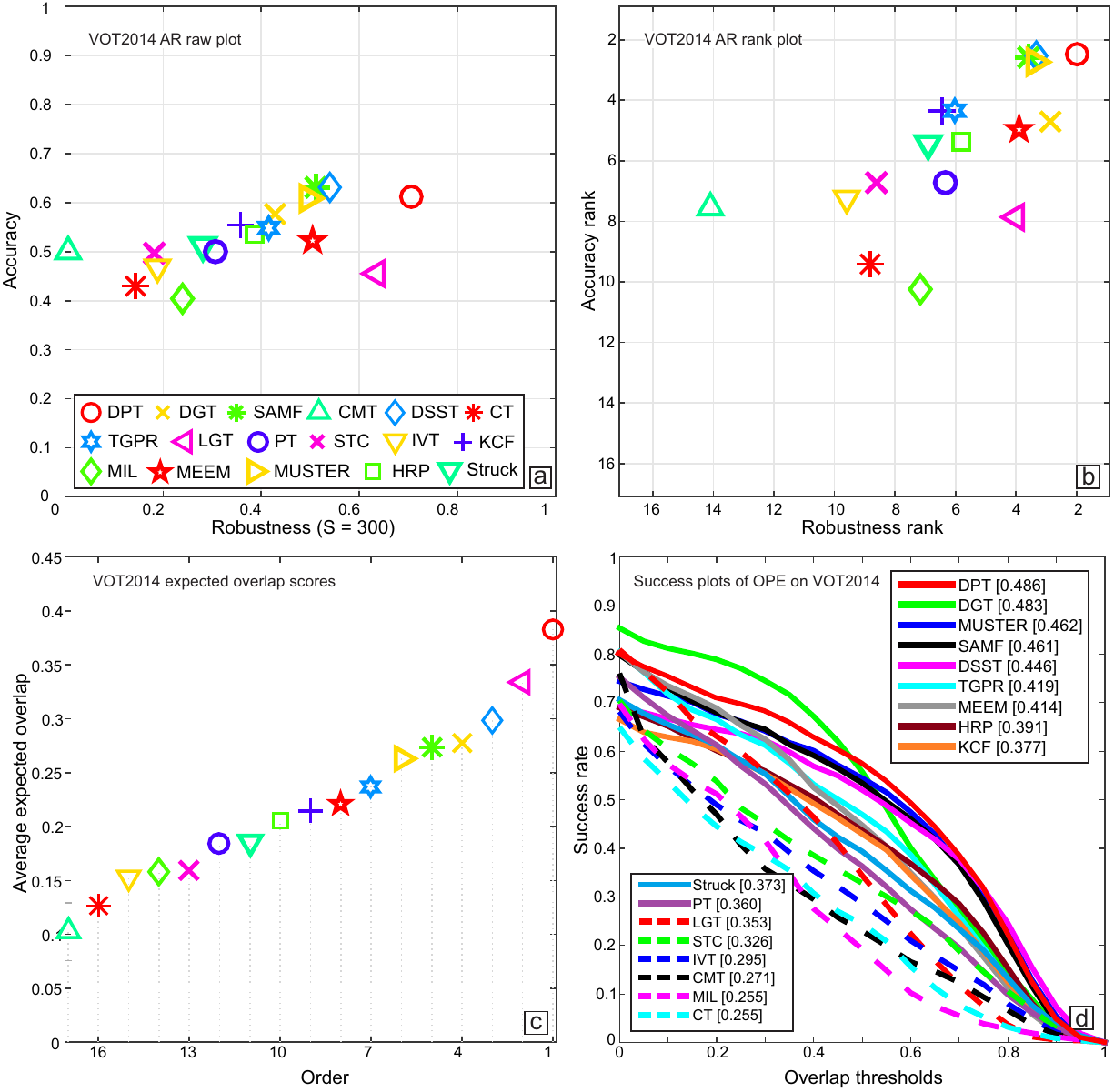}
\end{center}
   \caption{The VOT2014 AR raw (a), AR rank (b), expected average overlap (c) and the OTB success plot (d).}
\label{fig:baseline_sota_plots}
\end{figure}

The AR-raw, AR-rank and the expected average overlap plot of the VOT2014 reset-based experiment are shown in Figure~\ref{fig:baseline_sota_plots}(a,b,c). In terms of AR-raw and AR-rank plots, the DPT outperforms all trackers by being closest to the top-right part of the plots. The tracker exhibits excellent tradeoff between robustness and accuracy, attaining high accuracy during successful tracks and rarely fails. This is reflected in the average expected overlap measure, which ranks this tracker as a top performing tracker (Figure~\ref{fig:baseline_sota_plots}c and the last row in Table~\ref{tab:raw_attrib}). The DPT outperforms the best part-based tracker LGT~\cite{lgt_tpami2013} that applies a locally-connected constellation model and color segmentation by over $18\%$ and the winner of the VOT2014 challenge, the scale adaptive correlation filter DSST~\cite{danelljan2014accurate}, by $30\%$.

The VOT reset-based methodology resets the tracker after failure, but some trackers, like MUSTER~\cite{muster_cvpr15}, MEEM~\cite{meem_eccv14} and CMT~\cite{cmt_nabehay_cvpr2015} explicitly address target loss and implement mechanisms for target re-detection upon drifting. Although these are long-term capabilities and DPT is a short-term tracker that does not perform re-detection, we performed the no-reset OTB~\cite{otb_cvpr2010} experiment to gain further insights. The OTB~\cite{otb_cvpr2010} methodology reports the tracker overlap precision with respect to the intersection thresholds in a form a success plot (Figure~\ref{fig:baseline_sota_plots}d). The trackers are then ranked by the area under the curve (AUC) measure, which is equivalent to a no-reset average overlap~\cite{cehovin_tip2016}. The DPT outperforms the best baseline color-based superpixel short-term tracker DGT~\cite{dgt2014tip} and the long-term tracker MUSTER~\cite{muster_cvpr15}, which combines robust keypoint matching, correlation filter (DSST~\cite{danelljan2014accurate}), HoG and color features. The DPT also outperformed the recent state-of-the-art discriminative correlation filter-based trackers like DSST~\cite{danelljan2014accurate}, color-based SAMF~\cite{samf_eccv2014}, the recently proposed multi-snapshot online SVM-based MEEM~\cite{meem_eccv14} and the recent logistic regression tracker HRP~\cite{hrp_iccv15} tracker. The results conclusively show top global performance over the related state-of-the-art with respect to several performance measures and experimental setups.


\subsubsection{Per-attribute analysis}

Next we analyzed tracking performance with respect to the visual attributes. The VOT2014 benchmark provides a highly detailed per-frame annotation with the following attributes: \textit{camera motion}, \textit{illumination change}, \textit{occlusion}, \textit{size change} and \textit{motion change}. In addition to these, we manually annotated sequences that contained deformable targets by the \textit{deformation} attribute. If a frame did not contain any attribute or deforming target, it was annotated by an \textit{empty} attribute.

\begin{figure}
\begin{center}
\includegraphics[width=\linewidth]{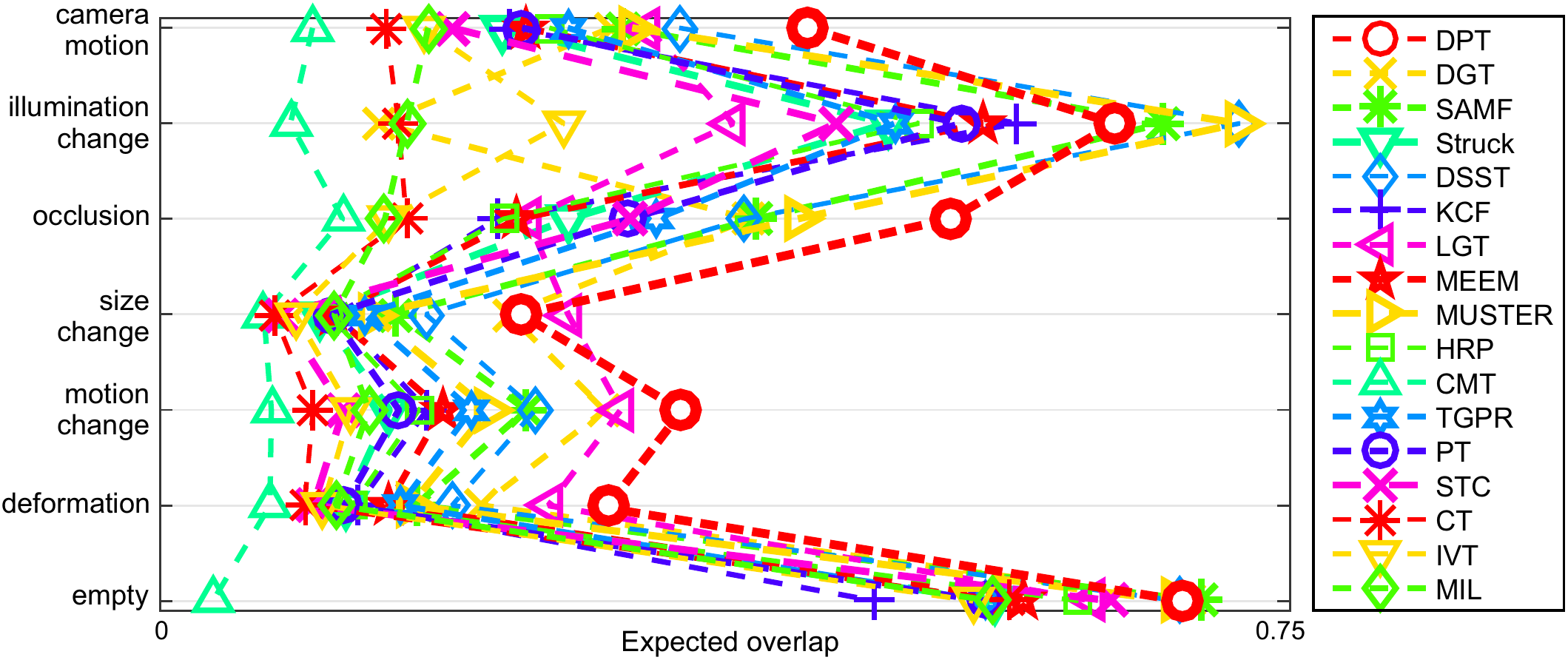}
\end{center}
   \caption{The expected average overlap with respect to the visual attributes on the VOT2014 dataset.}
\label{fig:expected-overlap-attributes}
\end{figure}

The tracking performance with respect to each attribute is shown in Figure~\ref{fig:expected-overlap-attributes} and Table~\ref{tab:raw_attrib}. The DPT outperforms all trackers on occlusion, camera motion, motion change and deformation and is among the top-performing trackers on illumination change, size change and empty. 
Note that the DPT outperformed all trackers that explicitly address target drift and partial occlusion, i.e., MUSTER~\cite{muster_cvpr15}, MEEM~\cite{meem_eccv14}, CMT~\cite{cmt_nabehay_cvpr2015}, Struck~\cite{hare_struck}. The DPT also outperforms top part-based trackers that address non-rigid deformations, i.e., LGT~\cite{lgt_tpami2013}, DGT~\cite{dgt2014tip}, PT~\cite{ruiyao_partbased_cvpr2013} and CMT~\cite{cmt_nabehay_cvpr2015}. These results indicate a balanced performance in that the DPT does not only excel at a given attribute but performs well over all visual attributes.

\begin{table*}[ht]
\vspace*{1em}
\caption{\label{tab:raw_attrib}The per-attribute expected average overlap, i.e., EAO measure, ($\Omega$), reset-based overlap (O) and number of failures (F) for the top 10 ranked trackers over 7 visual attributes: camera motion (CM), deformation (DE), empty (EM), illumination change (IC), motion change (MC), occlusion (OC), size change (SC). The arrows $\uparrow$ and $\downarrow$ indicate that ``higher is better" and ``lower is better", respectively.}
\centering
\scalebox{.6}{
\begin{tabular}{l|c c c|c c c|c c c|c c c|c c c|c c c|c c c|c c c|c c c|c c c}
\multirow{2}{*}{{ }} & \multicolumn{3}{ |c| }{{DPT}} & \multicolumn{3}{ |c| }{{LGT}~\cite{lgt_tpami2013}} & \multicolumn{3}{ |c| }{{DSST}~\cite{danelljan2014accurate}} & \multicolumn{3}{ |c| }{{DGT}~\cite{dgt2014tip}} & \multicolumn{3}{ |c| }{{SAMF}~\cite{samf_eccv2014}} & \multicolumn{3}{ |c| }{{MUSTER}~\cite{muster_cvpr15}} & \multicolumn{3}{ |c| }{{TGPR}~\cite{gaoECCV2014}} & \multicolumn{3}{ |c| }{{MEEM}~\cite{meem_eccv14}} & \multicolumn{3}{ |c| }{{KCF}~\cite{henriques2015tracking}} & \multicolumn{3}{ |c }{{HRP}~\cite{hrp_iccv15}}\\\hline
attr. & EAO$\uparrow$ & O$\uparrow$ & F$\downarrow$ & $\Omega\uparrow$ & O$\uparrow$ & F$\downarrow$ & $\Omega\uparrow$ & O$\uparrow$ & F$\downarrow$ & $\Omega\uparrow$ & O$\uparrow$ & F$\downarrow$ & $\Omega\uparrow$ & O$\uparrow$ & F$\downarrow$ & $\Omega\uparrow$ & O$\uparrow$ & F$\downarrow$ & $\Omega\uparrow$ & O$\uparrow$ & F$\downarrow$ & $\Omega\uparrow$ & O$\uparrow$ & F$\downarrow$ & $\Omega\uparrow$ & O$\uparrow$ & F$\downarrow$ & $\Omega\uparrow$ & O$\uparrow$ & F$\downarrow$ \\\hline \hline
{CM} & 0.43 & 0.64 & 12.00 & 0.32 & 0.44 & 15.20 & 0.34 & 0.66 & 20.00 & 0.30 & 0.56 & 19.00 & 0.31 & 0.65 & 24.00 & 0.31 & 0.63 & 22.00 & 0.27 & 0.57 & 27.27 & 0.24 & 0.53 & 25.00 & 0.23 & 0.57 & 34.00 & 0.26 & 0.58 & 30.00 \\
{DE} & 0.31 & 0.60 & 17.00 & 0.26 & 0.41 & 11.16 & 0.19 & 0.56 & 28.00 & 0.21 & 0.54 & 16.00 & 0.17 & 0.59 & 31.00 & 0.17 & 0.55 & 31.00 & 0.16 & 0.53 & 37.07 & 0.15 & 0.52 & 33.00 & 0.13 & 0.53 & 42.00 & 0.13 & 0.50 & 41.00 \\
{EM } & 0.68 & 0.49 & 0.00 & 0.62 & 0.51 & 0.00 & 0.68 & 0.54 & 0.00 & 0.68 & 0.67 & 0.00 & 0.69 & 0.56 & 0.00 & 0.67 & 0.51 & 0.00 & 0.55 & 0.41 & 0.00 & 0.57 & 0.47 & 0.00 & 0.47 & 0.56 & 0.00 & 0.61 & 0.26 & 0.00 \\
{IC } & 0.64 & 0.63 & 1.00 & 0.38 & 0.45 & 1.47 & 0.72 & 0.74 & 1.00 & 0.15 & 0.46 & 14.00 & 0.67 & 0.67 & 1.00 & 0.72 & 0.73 & 1.00 & 0.49 & 0.57 & 3.47 & 0.55 & 0.54 & 2.00 & 0.57 & 0.54 & 1.00 & 0.50 & 0.66 & 4.00 \\
{MC } & 0.35 & 0.63 & 14.00 & 0.31 & 0.46 & 10.47 & 0.25 & 0.64 & 24.00 & 0.30 & 0.58 & 14.00 & 0.24 & 0.66 & 25.00 & 0.22 & 0.64 & 26.00 & 0.21 & 0.55 & 30.20 & 0.19 & 0.53 & 24.00 & 0.18 & 0.57 & 34.00 & 0.17 & 0.60 & 35.00 \\
{OC } & 0.52 & 0.62 & 2.00 & 0.24 & 0.32 & 3.93 & 0.39 & 0.63 & 3.00 & 0.39 & 0.48 & 1.00 & 0.40 & 0.60 & 4.00 & 0.42 & 0.61 & 3.00 & 0.33 & 0.61 & 5.00 & 0.24 & 0.57 & 3.00 & 0.22 & 0.58 & 6.00 & 0.23 & 0.47 & 5.00 \\
{SC } & 0.24 & 0.54 & 12.00 & 0.27 & 0.43 & 7.40 & 0.18 & 0.52 & 15.00 & 0.23 & 0.57 & 6.00 & 0.16 & 0.56 & 18.00 & 0.14 & 0.53 & 19.00 & 0.14 & 0.47 & 21.20 & 0.11 & 0.46 & 15.00 & 0.12 & 0.47 & 27.00 & 0.11 & 0.50 & 27.00 \\\hline
{\average{Average} } & 0.39 & 0.61 & 11.97 & 0.33 & 0.44 & 11.16 & 0.30 & 0.62 & 19.28 & 0.28 & 0.56 & 14.31 & 0.27 & 0.63 & 21.76 & 0.26 & 0.61 & 21.39 & 0.24 & 0.54 & 25.76 & 0.22 & 0.52 & 22.04 & 0.21 & 0.55 & 30.24 & 0.20 & 0.55 & 29.13 \\
\end{tabular}
}
\end{table*}

\subsection{Performance on benchmarks} \label{sec:benchmarks}

\begin{figure}
\vspace*{1em}
\begin{center}
\includegraphics[width=\linewidth]{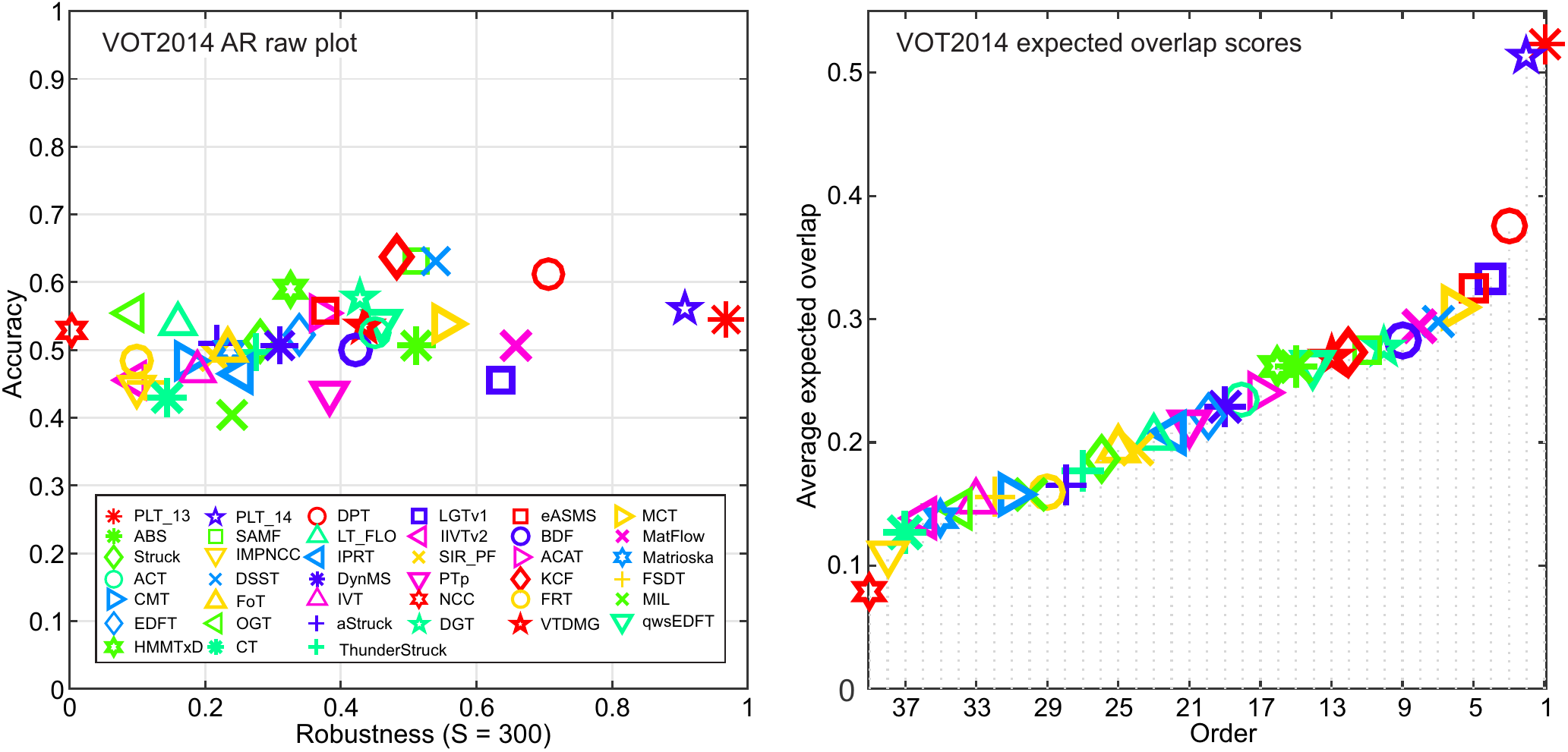}
\end{center}
   \caption{The AR raw plots and the expected average overlap accuracy measures for VOT2014 benchmark \cite{kristan_vot2014}. Please see~\cite{kristan_vot2014} for the tracker references.}
\label{fig:vot14-benchmark}
\end{figure}

For completeness of the analysis we have benchmarked the proposed tracker on the recent benchmarks. The DPT performance on the VOT2014 benchmark~\cite{kristan_vot2014} compared to the 38 trackers available in that benchmark is shown in Figure ~\ref{fig:vot14-benchmark}. The DPT excels in the reset-based accuracy, robustness as well as the expected average overlap accuracy measure and is ranked third, outperforming $92\%$ of the trackers on the benchmark. The two trackers that outperform the DPT are variants of the unpublished PLT tracker~\cite{kristan_vot2014}.

\begin{figure}[!ht]
\vspace*{1em}
\begin{center}
\includegraphics[width=\linewidth]{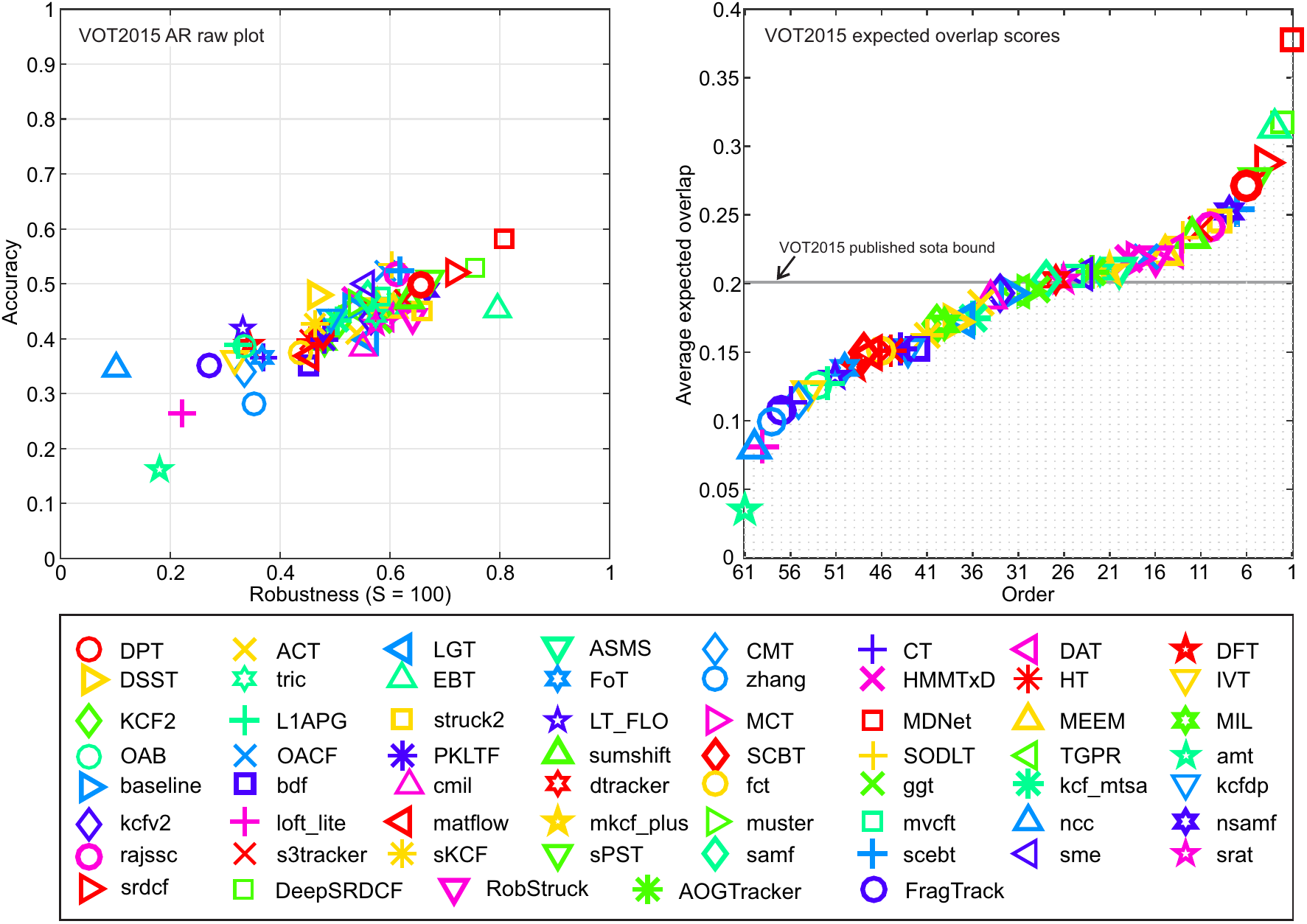}
\end{center}
   \caption{The AR raw plots and the expected average overlap accuracy measures for VOT2015 benchmark \cite{kristan_vot2015}. Please see~\cite{kristan_vot2015} for the tracker references.}
\label{fig:vot15-benchmark}
\end{figure}

The DPT performance on the most recent and challenging VOT2015 benchmark~\cite{kristan_vot2015} compared to the 60 trackers included in that benchmark are shown in Figure~\ref{fig:vot15-benchmark}. The tracker is ranked among the top $10\%$ of all trackers, outperforming 54 trackers (i.e., $90\%$ of the benchmark). The DPT outperforms all fifteen part-based trackers and fourteen correlation filter trackers, including the nSAMF, which is an improved version of~\cite{samf_eccv2014} that applies color as well as fusion with various models, and the recently published improved Struck~\cite{hare_pami2016} that applies additional features and performs remarkably well compared to the original version~\cite{hare_struck}. The VOT2015 provides a \textit{VOT2015 published sota bound} computed by averaging performance of trackers published in 2014/2015 in top computer vision conferences and journals. Any tracker with performance over this boundary is considered a state-of-the-art tracker according to VOT. The DPT is positioned well above this boundary and is considered a state-of-the-art according to the strict VOT2015 standards.

\begin{figure}[!ht]
\vspace*{1em}
\begin{center}
\includegraphics[width=0.9\linewidth]{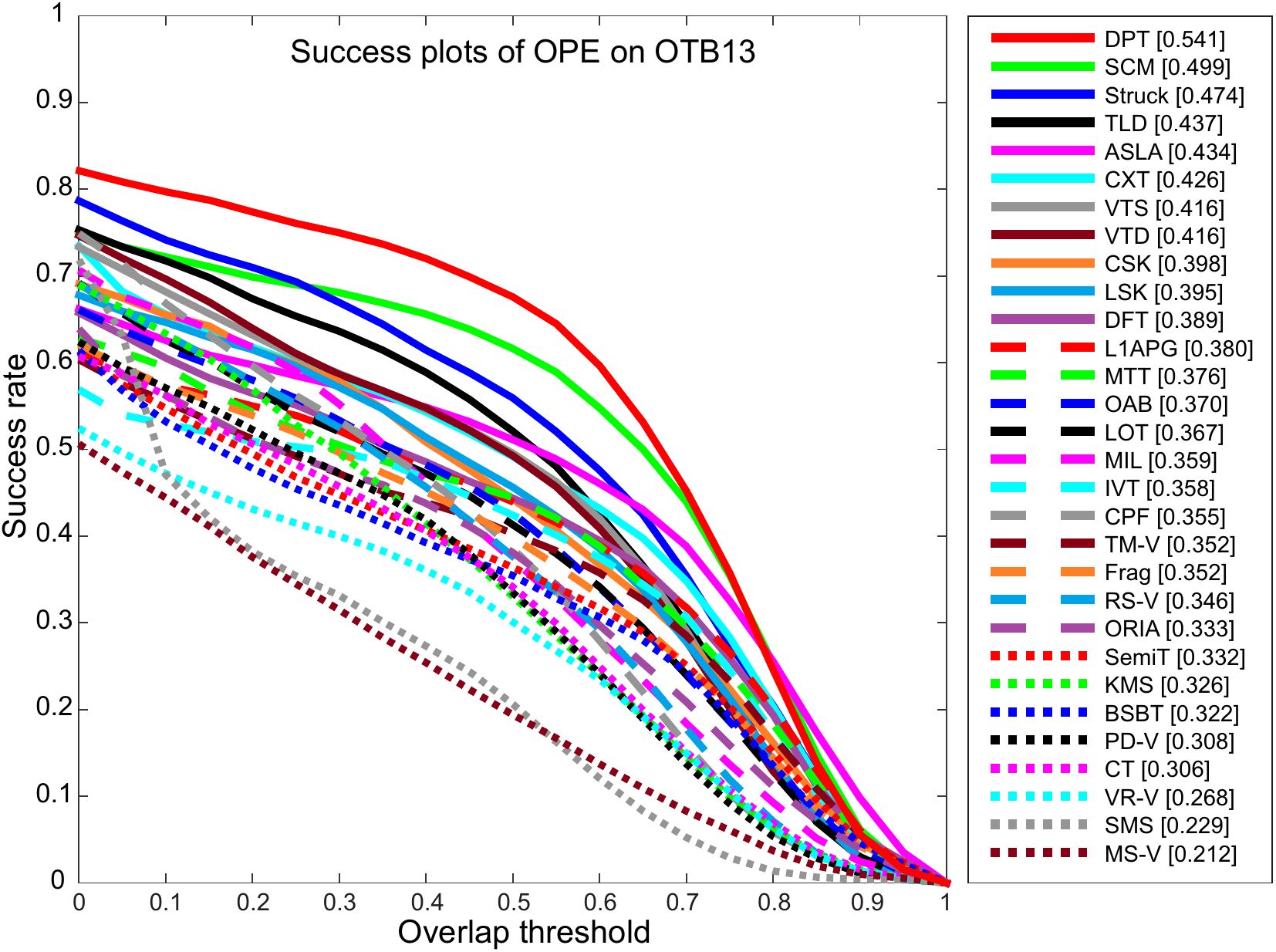}
\end{center}
   \caption{The OPE performance plot for the top trackers on the OTB benchmark~\cite{otb_cvpr2010}. Please see~\cite{otb_cvpr2010} for the tracker references.}
\label{fig:otb-success-plots}
\end{figure}

The DPT performance against 29 trackers available on the standard OTB~\cite{otb_cvpr2010} benchmark is shown in Figure~\ref{fig:otb-success-plots}. The DPT outperforms all trackers and is ranked top, exceeding the performance of the second-best tracker by over $8\%$.

\subsection{Qualitative analysis} \label{sec:qualitative}

\begin{figure}[h!]
\begin{center}
\includegraphics[width=\linewidth]{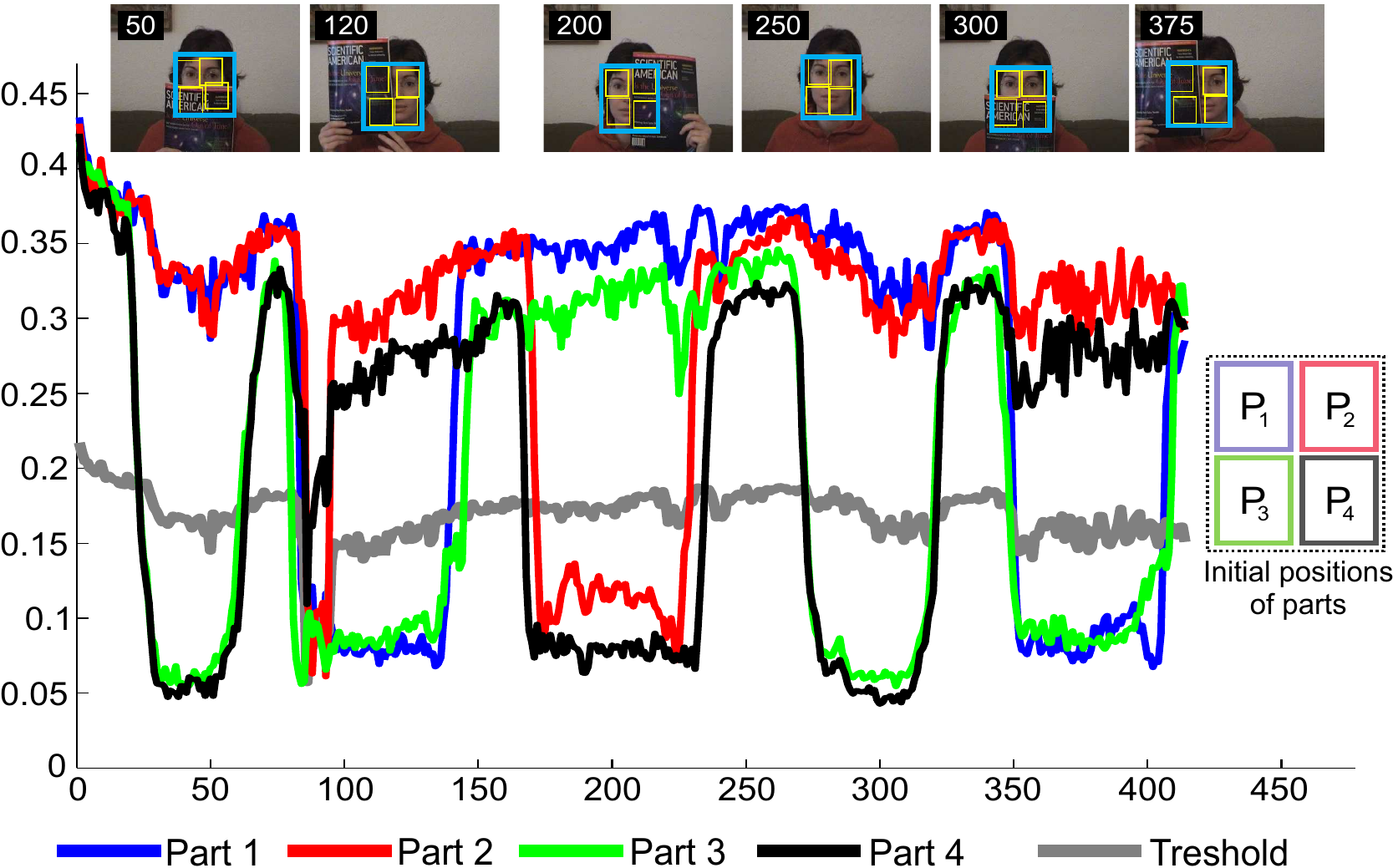}
\end{center}
   \caption{Qualitative tracking results of partially occluded object. A sketch of parts is shown on the right-hand side. Part weights are color-coded, with the update threshold shown in gray.}
\label{fig:face_occlusion}
\end{figure}

Qualitative analysis is provided for further insights. An experiment was performed to demonstrate the effectiveness of part adaptations during significant partial occlusions. The DPT was applied to a well-known sequence, in which the object (face) undergoes repetitive partial occlusions by a book (see Figure~\ref{fig:face_occlusion}). The DPT tracked the face without failures. Figure~\ref{fig:face_occlusion} shows images of the face taken from the sequence along with the graph of color-coded part weights ${w}_{t}^{(i)}$. The automatically computed adaptation threshold is shown in gray. Recall that part is updated if the weight exceeds this threshold (Section~\ref{sec:tracking}). Observe that partial occlusions are clearly identified by the weight graphs, resulting in drift prevention and successful tracking through partial occlusions.

\begin{figure*}[ht!]
\vspace*{1em}
\begin{center}
\includegraphics[width=\linewidth]{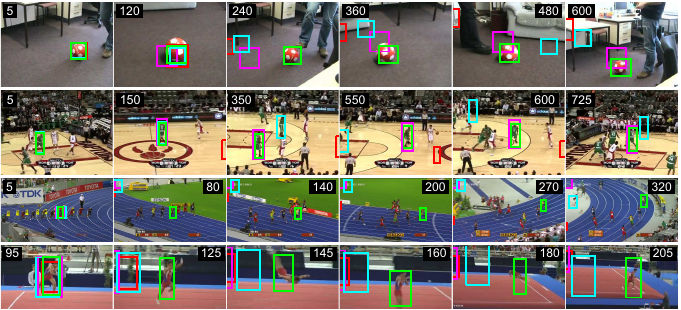}
\end{center}
   \caption{Qualitative comparative examples of tracking for DPT, DSST, MUSTER and Struck shown in green, red, magenta and cyan, respectively.}
\label{fig:qualitatComparison}
\end{figure*}

\begin{figure*}[!ht]
\vspace*{1em}
\begin{center}
\includegraphics[width=\linewidth]{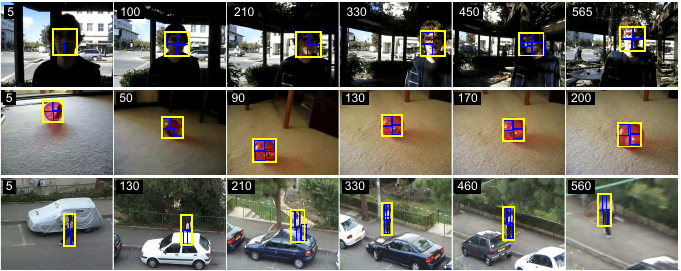}
\end{center}
   \caption{Qualitative examples of DPT tracker on three sequences. Tracking bounding box is visualized with yellow color and four parts on mid-level representation are shown in blue.}
\label{fig:ldp-frames}
\end{figure*}

Additional qualitative examples are provided in Figure~\ref{fig:qualitatComparison}. The first row in Figure~\ref{fig:qualitatComparison} shows performance on a non-deformable target with fast-varying local appearance. The DPT tracks the target throughout the sequence, while holistic correlation- and SVM-based trackers~\cite{danelljan2014accurate,muster_cvpr15,hare_struck} fail. The second, third and fourth row show tracking of deformable targets of various degrees of deformation. The fourth row shows tracking of a gymnast that drastically and rapidly changes the appearance. Note that the DPT comfortably tracks the target, while the related trackers fail. The first and second row in Figure~\ref{fig:ldp-frames} visualizes successful tracking performance on targets undergoing significant illumination changes. The third row shows tracking through several long-term partial occlusions. Again, the DPT successfully tracks the target even though the bottom part remains occluded for a large number of frames. The constellation model overcomes the occlusion and continues tracking during and after the occlusions.

\section{Conclusion} \label{sec:conclusion}
 
A new class of deformable parts trackers based on correlation filters is presented. The developed deformable parts model jointly treats the visual and geometric properties within a single formulation, resulting in a convex optimization problem. The parts appearance models are updated by online regression to result in Gaussian-like likelihood functions and the geometric constraints are modeled as a fully-connected spring system. We have shown that the dual representation of such a deformable parts model is an extended spring system and that minimization of the corresponding energy function leads to a MAP inference on the deformable parts model. A highly efficient optimization called iterated direct approach (IDA) is derived for this dual formulation. A deformable parts correlation filter tracker (DPT) is proposed that combines a coarse object representation with a mid-level constellation of deformable parts model in top-down localization and bottom-up updates.

The extensive analysis of the new spring-system optimization method IDA showed remarkable convergence and robustness properties. In particular, the IDA converges much faster than the conjugated gradient descent, is numerically more robust and scales very well with increasing the number of parts in the spring system. 
Our tracker was rigorously compared against the state-of-the-art with respect to several performance measures and experimental setups against sixteen state-of-the-art baselines. The DPT tracker outperforms the related state-of-the-art part-based trackers as well as state-of-the-art trackers that use a single appearance model, including the winner of the VOT2014 challenge and runs in real-time. Additional tests show that improvements come from the fully-connected constellation and the top-down/bottom-up combination of the coarse representation with the proposed deformable parts model. The DPT tracker was benchmarked on three recent highly challenging benchmarks against 38 trackers on VOT2014~\cite{kristan_vot2014} benchmark, 60 trackers on VOT2015~\cite{kristan_vot2015} benchmark and 29 trackers on the OTB~\cite{otb_cvpr2010} benchmark. The DPT attained a state-of-the-art performance on all benchmarks. Note that, since five KCFs~\cite{henriques2015tracking} are used in DPT, the speed reduction is approximately five times compared to the baseline KCF. But the boost in performance is significant. 
The DPT reduces the failures compared to the baseline KCF by nearly $60\%$, the expected average overlap is increased by over $80\%$ and the OTB average overlap is increased by approximately $30\%$ while still attaining real-time performance.

The proposed deformable parts model is highly extendable. The dual formulation of the deformable constellation and the proposed optimizer are generally applicable as stand-alone solvers for deformable parts models. The appearance models on parts can be potentially replaced with other discriminative or generative models or augmented to obtain a constellation of parts based on various features like key-points and parts of different shapes. The part-based models like flocks of features~\cite{vojir_fot_2014}, key-point-based~\cite{Pernici2013,cmt_nabehay_cvpr2015} and superpixel-based~\cite{dgt2014tip} typically use more parts than the tracker presented in this paper. Our analysis shows that the proposed optimization of the deformation model scales well with the number of parts, and could be potentially used in these trackers as a deformation model. Parts could also be replaced with scale-adaptive parts, which could further improve scale adaptation of the whole tracker. Alternatively, saliency regions could be used to improve localization. One way to introduce the saliency is at the coarse layer and another to apply it at the parts localization. Since the model is fully probabilistic, it can be readily integrated with probabilistic dynamic models. These will be the topics of our future work.





\ifCLASSOPTIONcaptionsoff
  \newpage
\fi




\bibliographystyle{IEEEtran}
\bibliography{mybib}

\begin{thebibliography}{10}
\providecommand{\url}[1]{#1}
\csname url@samestyle\endcsname
\providecommand{\newblock}{\relax}
\providecommand{\bibinfo}[2]{#2}
\providecommand{\BIBentrySTDinterwordspacing}{\spaceskip=0pt\relax}
\providecommand{\BIBentryALTinterwordstretchfactor}{4}
\providecommand{\BIBentryALTinterwordspacing}{\spaceskip=\fontdimen2\font plus
\BIBentryALTinterwordstretchfactor\fontdimen3\font minus
  \fontdimen4\font\relax}
\providecommand{\BIBforeignlanguage}[2]{{%
\expandafter\ifx\csname l@#1\endcsname\relax
\typeout{** WARNING: IEEEtran.bst: No hyphenation pattern has been}%
\typeout{** loaded for the language `#1'. Using the pattern for}%
\typeout{** the default language instead.}%
\else
\language=\csname l@#1\endcsname
\fi
#2}}
\providecommand{\BIBdecl}{\relax}
\BIBdecl

\bibitem{otb_cvpr2010}
Y.~Wu, J.~Lim, and M.-H. Yang, ``Online object tracking: A benchmark,'' in
  \emph{Comp. Vis. Patt. Recognition}, 2013, pp. 2411-- 2418.

\bibitem{alov_pami2014}
A.~Smeulders, D.~Chu, R.~Cucchiara, S.~Calderara, A.~Dehghan, and M.~Shah,
  ``Visual tracking: An experimental survey,'' \emph{IEEE Trans. Pattern Anal.
  Mach. Intell.}, vol.~36, no.~7, pp. 1442--1468, July 2014.

\bibitem{kristan_vot2013}
M.~Kristan, R.~Pflugfelder, A.~Leonardis, J.~Matas, F.~Porikli, L.~Čehovin,
  G.~Nebehay, G.~Fernandez, and T.~e.~a. Vojir, ``The visual object tracking
  vot2013 challenge results,'' in \emph{Vis. Obj. Track. Challenge VOT2013, In
  conjunction with ICCV2013}, Dec 2013, pp. 98--111.

\bibitem{kristan_vot2014}
M.~Kristan, R.~Pflugfelder, A.~Leonardis, J.~Matas, L.~\v{C}ehovin, G.~Nebehay,
  T.~Vojir, and G.~et~al. Fernandez, ``The visual object tracking vot2014
  challenge results,'' in \emph{Proc. European Conf. Computer Vision}, 2014,
  pp. 191--217.

\bibitem{hare_struck}
S.~Hare, A.~Saffari, and P.~H.~S. Torr, ``Struck: Structured output tracking
  with kernels,'' in \emph{Int. Conf. Computer Vision}.\hskip 1em plus 0.5em
  minus 0.4em\relax Washington, DC, USA: IEEE Computer Society, 2011, pp.
  263--270.

\bibitem{babenko_mil}
B.~Babenko, M.-H. Yang, and S.~Belongie, ``Robust object tracking with online
  multiple instance learning,'' \emph{IEEE Trans. Pattern Anal. Mach. Intell.},
  vol.~33, no.~8, pp. 1619--1632, Aug. 2011.

\bibitem{grabner_oab}
H.~Grabner, M.~Grabner, and H.~Bischof, ``Real-time tracking via on-line
  boosting,'' in \emph{Proc. British Machine Vision Conference}, vol.~1, 2006,
  pp. 47--56.

\bibitem{bolme2010visual}
D.~S. Bolme, J.~R. Beveridge, B.~A. Draper, and Y.~M. Lui, ``Visual object
  tracking using adaptive correlation filters,'' in \emph{Comp. Vis. Patt.
  Recognition}.\hskip 1em plus 0.5em minus 0.4em\relax IEEE, 2010, pp.
  2544--2550.

\bibitem{danelljan2014accurate}
M.~Danelljan, G.~H{\"a}ger, F.~S. Khan, and M.~Felsberg, ``Accurate scale
  estimation for robust visual tracking,'' in \emph{Proc. British Machine
  Vision Conference}, 2014, pp. 1--11.

\bibitem{samf_eccv2014}
Y.~Li and J.~Zhu, ``A scale adaptive kernel correlation filter tracker with
  feature integration,'' in \emph{Proc. European Conf. Computer Vision}, 2014,
  pp. 254--265.

\bibitem{henriques2015tracking}
J.~F. Henriques, R.~Caseiro, P.~Martins, and J.~Batista, ``High-speed tracking
  with kernelized correlation filters,'' \emph{IEEE Trans. Pattern Anal. Mach.
  Intell.}, vol.~37, no.~3, pp. 583--596, 2014.

\bibitem{comanichu_kernel_pami2003}
D.~Comaniciu, V.~Ramesh, and P.~Meer, ``Kernel-based object tracking,''
  \emph{IEEE Trans. Pattern Anal. Mach. Intell.}, vol.~25, no.~5, pp. 564--577,
  May 2003.

\bibitem{lgt_tpami2013}
L.~Čehovin, M.~Kristan, and A.~Leonardis, ``Robust visual tracking using an
  adaptive coupled-layer visual model,'' \emph{IEEE Trans. Pattern Anal. Mach.
  Intell.}, vol.~35, no.~4, pp. 941--953, Apr. 2013.

\bibitem{kwon_tracking_sampling_tpami2014}
J.~Kwon and K.~M. Lee, ``Tracking by sampling and integrating multiple
  trackers,'' \emph{IEEE Trans. Pattern Anal. Mach. Intell.}, vol.~36, no.~7,
  pp. 1428--1441, July 2014.

\bibitem{Maresca2013}
M.~E. Maresca and A.~Petrosino, ``Matrioska: A multi-level approach to fast
  tracking by learning,'' in \emph{Proc. Int. Conf. Image Analysis and
  Processing}, 2013, pp. 419--428.

\bibitem{pr2011_artner}
N.~M. Artner, A.~Ion, and W.~G. Kropatsch, ``Multi-scale 2d tracking of
  articulated objects using hierarchical spring systems,'' \emph{Patt.
  Recogn.}, vol.~44, no.~4, pp. 800--810, 2011.

\bibitem{dgt2014tip}
Z.~Cai, L.~Wen, Z.~Lei, N.~Vasconcelos, and S.~Li, ``Robust deformable and
  occluded object tracking with dynamic graph,'' \emph{IEEE Trans. Image
  Proc.}, vol.~23, no.~12, pp. 5497 -- 5509, 2014.

\bibitem{ruiyao_partbased_cvpr2013}
R.~Yao, Q.~Shi, C.~Shen, Y.~Zhang, and A.~van~den Hengel, ``Part-based visual
  tracking with online latent structural learning,'' in \emph{Comp. Vis. Patt.
  Recognition}, June 2013, pp. 2363--2370.

\bibitem{godec2013cviu}
M.~Godec, P.~M. Roth, and H.~Bischof, ``Hough-based tracking of non-rigid
  objects.'' \emph{Comp. Vis. Image Understanding}, vol. 117, no.~10, pp.
  1245--1256, 2013.

\bibitem{part_context_bmvc2014}
G.~Zhu, J.~Wang, C.~Zhao, and H.~Lu, ``Part context learning for visual
  tracking,'' in \emph{Proc. British Machine Vision Conference}, 2014, pp.
  1--12.

\bibitem{Collins2005}
R.~T. Collins, X.~Liu, and M.~Lordeanu, ``Online selection of discriminative
  tracking features,'' \emph{IEEE Trans. Pattern Anal. Mach. Intell.}, vol.~27,
  no.~10, pp. 1631--1643, 2005.

\bibitem{ross_ivt}
D.~A. Ross, J.~Lim, R.-S. Lin, and M.-H. Yang, ``Incremental learning for
  robust visual tracking,'' \emph{Int. J. Comput. Vision}, vol.~77, no. 1-3,
  pp. 125--141, May 2008.

\bibitem{zhang_ct}
K.~Zhang, L.~Zhang, and M.-H. Yang, ``Real-time compressive tracking,'' in
  \emph{Proc. European Conf. Computer Vision}, 2012, pp. 864--877.

\bibitem{mei_tpami2011}
X.~Mei and H.~Ling, ``Robust visual tracking and vehicle classification via
  sparse representation,'' \emph{IEEE Trans. Pattern Anal. Mach. Intell.},
  vol.~33, no.~11, pp. 2259--2272, Nov 2011.

\bibitem{hong_multitask_view_iccv2013}
Z.~Hong, X.~Mei, D.~Prokhorov, and D.~Tao, ``Tracking via robust multi-task
  multi-view joint sparse representation,'' in \emph{Int. Conf. Computer
  Vision}, Dec 2013, pp. 649--656.

\bibitem{gaoECCV2014}
J.~Gao, H.~Ling, W.~Hu, and J.~Xing, ``Transfer learning based visual tracking
  with gaussian processes regression,'' in \emph{Proc. European Conf. Computer
  Vision}, vol. 8691, 2014, pp. 188--203.

\bibitem{avidan_svm_tracking}
S.~Avidan, ``Support vector tracking,'' \emph{IEEE Trans. Pattern Anal. Mach.
  Intell.}, vol.~26, no.~8, pp. 1064--1072, Aug 2004.

\bibitem{posseger_color_cvpr15}
H.~Possegger, T.~Mauthner, and H.~Bischof, ``In defense of color-based
  model-free tracking,'' in \emph{Comp. Vis. Patt. Recognition}, 2015, pp.
  2113--2120.

\bibitem{Naidu1974}
P.~Naidu, ``Improved optical character recognition by matched filtering,''
  \emph{Optics Communications}, vol.~12, no.~3, pp. 287--289, 1974.

\bibitem{zhang_2015_robust}
M.~Zhang, J.~Xing, J.~Gao, and W.~Hu, ``Robust visual tracking using joint
  scale-spatial correlation filters,'' in \emph{Proc. Int. Conf. Image
  Processing}.\hskip 1em plus 0.5em minus 0.4em\relax IEEE, 2015, pp.
  1468--1472.

\bibitem{li_2015_reliable}
Y.~Li, J.~Zhu, and S.~C. Hoi, ``Reliable patch trackers: Robust visual tracking
  by exploiting reliable patches,'' in \emph{Comp. Vis. Patt. Recognition},
  2015, pp. 353--361.

\bibitem{zhang_stc_eccv2014}
K.~Zhang, L.~Zhang, Q.~Liu, D.~Zhang, and M.-H. Yang,
  ``\BIBforeignlanguage{English}{Fast visual tracking via dense spatio-temporal
  context learning},'' in \emph{\BIBforeignlanguage{English}{Proc. European
  Conf. Computer Vision}}.\hskip 1em plus 0.5em minus 0.4em\relax Springer
  International Publishing, 2014, pp. 127--141.

\bibitem{muster_cvpr15}
Z.~Hong, Z.~Chen, C.~Wang, X.~Mei, D.~Prokhorov, and D.~Tao, ``Multi-store
  tracker (muster): A cognitive psychology inspired approach to object
  tracking,'' in \emph{Comp. Vis. Patt. Recognition}, 2015, pp. 749--758.

\bibitem{Hoey2006}
J.~Hoey, ``Tracking using flocks of features, with application to assisted
  handwashing,'' in \emph{Proc. British Machine Vision Conference}, vol.~1,
  2006, pp. 367--376.

\bibitem{vojir_fot_2014}
T.~Vojir and J.~Matas, ``The enhanced flock of trackers,'' in
  \emph{Registration and Recognition in Images and Videos}, ser. Studies in
  Computational Intelligence.\hskip 1em plus 0.5em minus 0.4em\relax Springer
  Berlin Heidelberg, 2014, vol. 532, pp. 113--136.

\bibitem{Martinez2008}
B.~Martinez and X.~Binefa, ``Piecewise affine kernel tracking for non-planar
  targets,'' \emph{Patt. Recogn.}, vol.~41, no.~12, pp. 3682--3691, 2008.

\bibitem{Pernici2013}
F.~Pernici and A.~Del~Bimbo, ``Object tracking by oversampling local
  features,'' \emph{IEEE Trans. Pattern Anal. Mach. Intell.}, vol.~36, no.~12,
  pp. 2538--2551, 2013.

\bibitem{yangreal}
X.~Yang, Q.~Xiao, S.~Wang, and P.~Liu, ``Real-time tracking via deformable
  structure regression learning,'' in \emph{Proc. Int. Conf. Pattern
  Recognition}, 2014, pp. 2179--2184.

\bibitem{Duffner2013}
S.~Duffner and C.~Garcia, ``{PixelTrack: a fast adaptive algorithm for tracking
  non-rigid objects},'' in \emph{Int. Conf. Computer Vision}, 2013, pp.
  2480--2487.

\bibitem{duan_2012_group}
G.~Duan, H.~Ai, S.~Cao, and S.~Lao, ``Group tracking: exploring mutual
  relations for multiple object tracking,'' in \emph{Proc. European Conf.
  Computer Vision}.\hskip 1em plus 0.5em minus 0.4em\relax Springer, 2012, pp.
  129--143.

\bibitem{kristan_vot2015}
M.~Kristan, J.~Matas, A.~Leonardis, M.~Felsberg, L.~\v{C}ehovin, G.~Fernandez,
  T.~Vojir, G.~H\"{a}ger, G.~Nebehay, and R.~et~al. Pflugfelder, ``The visual
  object tracking vot2015 challenge results,'' in \emph{Int. Conf. Computer
  Vision}, 2015.

\bibitem{danelljan_color_cf_cvpr14}
M.~Danelljan, F.~Shahbaz~Khan, M.~Felsberg, and J.~van~de Weijer, ``Adaptive
  color attributes for real-time visual tracking,'' in \emph{Comp. Vis. Patt.
  Recognition}, 2014, pp. 1090--1097.

\bibitem{kristan_segmentation_accv14}
M.~Kristan, J.~Perš, V.~Sulič, and S.~Kovačič, ``A graphical model for
  rapid obstacle image-map estimation from unmanned surface vehicles,'' in
  \emph{Proc. Asian Conf. Computer Vision}, 2014, pp. 391--406.

\bibitem{diplaros_genmodel}
A.~Diplaros, N.~Vlassis, and T.~Gevers, ``A spatially constrained generative
  model and an em algorithm for image segmentation,'' \emph{IEEE Trans. Neural
  Networks}, vol.~18, no.~3, pp. 798 -- 808, 2007.

\bibitem{dalal_triggs_hog}
N.~Dalal and B.~Triggs, ``Histograms of oriented gradients for human
  detection,'' in \emph{Comp. Vis. Patt. Recognition}, vol.~1, June 2005, pp.
  886--893.

\bibitem{linear_nonlinear_optimization}
I.~Griva, S.~G. Nash, and A.~Sofer, \emph{Linear and Nonlinear Optimization,
  Second Edition}.\hskip 1em plus 0.5em minus 0.4em\relax Siam, 2009.

\bibitem{Kristan2015}
M.~Kristan, J.~Matas, A.~Leonardis, T.~Vojir, R.~Pflugfelder, G.~Fernandez,
  G.~Nebehay, F.~Porikli, and L.~Cehovin, ``A novel performance evaluation
  methodology for single-target trackers,'' \emph{IEEE Trans. Pattern Anal.
  Mach. Intell.}, 2016.

\bibitem{cmt_nabehay_cvpr2015}
G.~Nebehay and R.~Pflugfelder, ``Clustering of static-adaptive correspondences
  for deformable object tracking,'' in \emph{Comp. Vis. Patt. Recognition},
  2015, pp. 2784--2791.

\bibitem{meem_eccv14}
J.~Zhang, S.~Ma, and S.~Sclaroff, ``{MEEM:} robust tracking via multiple
  experts using entropy minimization,'' in \emph{Proc. European Conf. Computer
  Vision}, 2014, pp. 188--203.

\bibitem{hrp_iccv15}
N.~Wang, J.~Shi, D.-Y. Yeung, and J.~Jia, ``Understanding and diagnosing visual
  tracking systems,'' in \emph{Int. Conf. Computer Vision}, 2015.

\bibitem{cehovin_tip2016}
L.~Čehovin, A.~Leonardis, and M.~Kristan, ``Visual object tracking performance
  measures revisited,'' \emph{IEEE Trans. Image Proc.}, vol.~25, no.~3, pp.
  1261 -- 1274, 2016.

\bibitem{hare_pami2016}
S.~Hare, S.~Golodetz, A.~Saffari, V.~Vineet, M.~Cheng, S.~Hicks, and P.~Torr,
  ``Struck: Structured output tracking with kernels,'' \emph{IEEE Trans.
  Pattern Anal. Mach. Intell.}, 2016.

\end{thebibliography}

%


%

\begin{IEEEbiography}[{\includegraphics[width=1in,height=1.25in,clip,keepaspectratio]{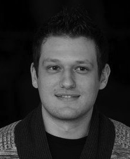}}]{Alan Lukežič} 
received the Dipl.ing. and M.Sc. degrees at the Faculty of Computer and Information Science, University of Ljubljana, Slovenia in 2012 and 2015, respectively. He is currently a researcher at the Visual Cognitive Systems Laboratory, Faculty of Computer and Information Science, University of Ljubljana, Slovenia as a researcher. His research interests include computer vision, data mining and machine learning.
\end{IEEEbiography}\vspace{-1cm}

\begin{IEEEbiography}[{\includegraphics[width=1in,height=1.25in,clip,keepaspectratio]{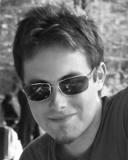}}]{Luka Čehovin} received his Ph.D from the Faculty of Computer and Information
Science, University of Ljubljana, Slovenia in 2015.
Currently he is working at the Visual Cognitive Systems Laboratory,
Faculty of Computer and Information Science, University of Ljubljana,
Slovenia as a teaching assistant and a researcher.
His research interests include computer vision, HCI, distributed
intelligence and web-mobile technologies.
\end{IEEEbiography}\vspace{-1cm}

\begin{IEEEbiography}[{\includegraphics[width=1in,height=1.25in,clip,keepaspectratio]{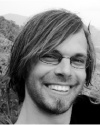}}]{Matej Kristan} received a Ph.D from the Faculty of Electrical Engineering, University
of Ljubljana in 2008. He is an Assistant Professor at the ViCoS
Laboratory at the Faculty of Computer and Information Science and at the
Faculty of Electrical Engineering, University of Ljubljana. His research
interests include probabilistic methods for computer vision with focus
on visual tracking, dynamic models, online learning, object detection
and vision for mobile robotics.
\end{IEEEbiography}\vspace{-1cm}



\end{document}